\documentclass[10pt,twocolumn,letterpaper]{article}

\usepackage[pagenumbers]{cvpr} % To force page numbers, e.g. for an arXiv version

\definecolor{cvprblue}{rgb}{0.21,0.49,0.74}
\usepackage[pagebackref,breaklinks,colorlinks,allcolors=cvprblue]{hyperref}
\usepackage{url}
\usepackage{comment}
\usepackage{amsmath,amssymb} 
\usepackage{multirow}
\usepackage{booktabs}
\usepackage{subcaption}
\usepackage{bm}
\usepackage{array}
\usepackage{xcolor}
\usepackage{graphicx}
\usepackage{enumitem}
\usepackage{makecell}
\usepackage{setspace}
\usepackage[symbol,hang,flushmargin]{footmisc}
\usepackage{wrapfig}
\usepackage{sidecap}
\usepackage{pgfplots}
\pgfplotsset{compat=1.18}
\usepackage[accsupp]{axessibility}
\usepackage[titletoc,toc,title]{appendix}

\def\paperID{13897} % *** Enter the Paper ID here
\def\confName{CVPR}
\def\confYear{2025}

\title{Enhancing Virtual Try-On with Synthetic Pairs and Error-Aware Noise Scheduling}

\author{Nannan Li\\
Boston University\\
{\tt\small nnli@bu.edu}
\and
Kevin J. Shih\\
NVIDIA\\
\and
Bryan A. Plummer \\
Boston University\\
{\tt\small bplum@bu.edu}
}

\newcommand{\Ours}{EARSB}
\newcommand{\Sch}{Schr\"odinger}
\renewcommand{\paragraph}[1]{ \noindent \textbf{#1.}}
\AtBeginEnvironment{equation}{
\setlength{\abovedisplayskip}{3.5pt}
\setlength{\belowdisplayskip}{3.5pt}
}
\AtBeginEnvironment{equation*}{
\setlength{\abovedisplayskip}{3.5pt}
\setlength{\belowdisplayskip}{3.5pt}
}
\AtBeginEnvironment{gather}{
\setlength{\abovedisplayskip}{3pt}
\setlength{\belowdisplayskip}{3pt}
}

\begin{document}
\maketitle
\vspace{-0.2cm}
% \input{sec/0_abstract}    
% \input{sec/1_intro}
% \input{sec/2_formatting}
% \input{sec/3_finalcopy}

% \twocolumn[{%
% \renewcommand\twocolumn[1][]{#1}%
% \maketitle
% \begin{center}
% \vspace{-0.55cm}
% \includegraphics[width=0.9\textwidth]{figures/fig1.png}
% \vspace{-1em}
%     \captionof{figure}{Refinement examples of our proposed Error-Aware Refinement \Sch{} Bridge \Ours{}. From left to right is the source human image, the garment image, the initial try-on image generated by an existing try-on model \citep{stableviton,sd-viton}, the image refined by our \Ours{}, and the image further refined by \Ours{} with \emph{synthetic data augmentation} in training.}
%     \label{fig:motiv}
% \end{center}
% }]
%diffusion model \citep{stableviton} (top row) or GAN \citep{sd-viton} (bottom row)

\begin{abstract}
Given an isolated garment image in a canonical product view and a separate image of a person, the virtual try-on task aims to generate a new image of the person wearing the target garment.
Prior virtual try-on works face two major challenges in achieving this goal: a) the paired (human, garment) training data has limited availability; b) generating textures on the human that perfectly match that of the prompted garment is difficult, often resulting in distorted text and faded textures. 
Our work explores ways to tackle these issues through both synthetic data as well as model refinement. We introduce a garment extraction model that generates (human, synthetic garment) pairs from a single image of a clothed individual. The synthetic pairs can then be used to augment the training of virtual try-on. We also propose an Error-Aware Refinement-based \Sch{} Bridge (\Ours{}) that surgically targets localized generation errors for correcting the output of a base virtual try-on model. To identify likely errors, we propose a weakly-supervised error classifier that localizes regions for refinement, subsequently augmenting the Schr\"odinger Bridge's noise schedule with its confidence heatmap. Experiments on VITON-HD and DressCode-Upper demonstrate that our synthetic data augmentation enhances the performance of prior work, while \Ours{} improves the overall image quality. In user studies, our model is preferred by the users in an average of 59\% of cases. Code is available at \href{https://github.com/NannanLi999/earsb.git}{this link}. 
\end{abstract}

\section{Introduction}

\begin{figure}
    \centering
    \includegraphics[width=\linewidth]{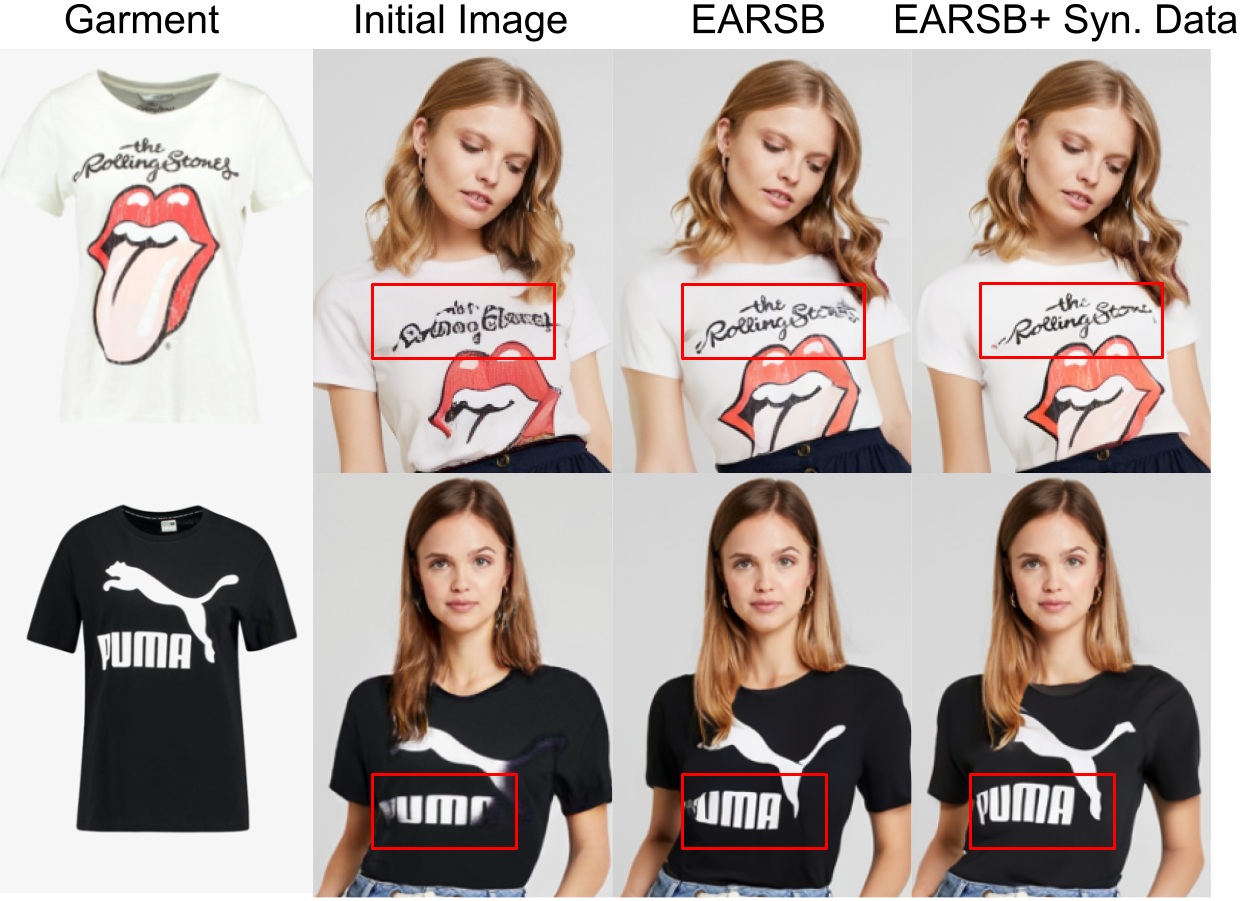}
    \caption{Example of our proposed Error-Aware Refinement \Sch{} Bridge (\Ours). \Ours{} can refine the artifacts (marked by bounding boxes) in an initial image generated by an existing try-on model. The initial image is generated by \citep{stableviton} in the top row and by \citep{sd-viton}  in the bottom row. + Syn. Data in the last column strengthens the refinement with the proposed synthetic data augmentation in training.}
    \label{fig:motiv}
    \vspace{-1em}
\end{figure}

Virtual try-on aims to generate a photorealistic image of a target person wearing a prompted product-view garment \citep{keypoints-tryon, tryondiffusion, tpd}. It allows users to visualize how garments would fit and appear on their bodies without the need for physical trials. 
While recent methods have made significant strides in this field \citep{sd-viton,gp-vton,tpd,stableviton}, noticeable artifacts such as text distortion and faded textures persist in generated images. For example, as illustrated in the second row of \cref{fig:motiv}, the logo and the text on the t-shirt noticeably fade away in the initial image generated by a prior try-on model \citep{sd-viton}. These imperfections stem from two primary challenges in virtual try-on: limited data availability and the complexity of accurate garment texture deformation. To address these issues, we propose a two-pronged approach: augmenting training data through cost-effective synthetic data generation, and surgically targeting known generation artifacts using our proposed Error-Aware Refinement-based Schr\"odinger Bridge (\Ours{}).

At a minimum, the training data of virtual try-on requires paired (human, product-view garment) images. The product-view garment image is a canonical, front-facing view of the clothing with a clean background. A substantial amount of data is needed to capture the combinatorial space comprising all possible human poses, skin tones, viewing angles, and their respective physical interactions with fabric textures, shapes, letterings, and other material properties. Unfortunately, these images are generally available only on copyright-protected product webpages and, therefore, are not readily available for use. 
To mitigate this issue, we propose to augment training with synthetic data generated from the easier symmetric human-to-garment task, wherein we train a garment-extraction model to extract a canonical product-view garment image from an image of a clothed person. This will allow us to create synthetic \emph{paired} training data from unpaired datasets \citep{deepfashion,sshq,upt}. Our results demonstrate that incorporating the more readily available synthetic training pairs can improve image generation quality in the virtual try-on task.

In addition to addressing the data scarcity issue, we aim to construct a refinement model that can make localized adjustments to a weaker model's generation results. Our approach draws inspiration from classical boosting approaches where every model in a cascade of models targets the shortcomings of the preceding models. We are interested in a targeted refinement approach for two main reasons: it allows a training objective that is focused solely on fixing specific errors, and potentially saves computation when initial predictions are sufficiently good.
%it allows us to construct a training objective that is solely focused on fixing specific generation errors. Second, we should be able to save on computation if the initial prediction is sufficiently good.

Two components are necessary to achieve such a pipeline: a classifier for identifying localized generation errors, and a refinement model that can re-synthesize content specifically in these localized regions. We found that an effective Weakly-Supervised error Classifier (WSC) can be constructed with just a few hours of manual labeling of generation errors. Another benefit of this approach is that it can be easily tailored for the errors of a specific model that produces images with artifacts. The resulting WSC will produce an error map highlighting low-quality regions. Subsequently, we adopt an Image-to-Image Schrodinger Bridge (I$^2$SB) \citep{i2sb} to learn the refinement of these regions in the generated images. While typical diffusion models map from noise to data, I$^2$SB constructs a \Sch{} Bridge (SB) that allows us to map from data to data, or in our setup, generations with artifacts to ground truth images. In addition, we introduce an \emph{adaptive} noise schedule to direct the SB process to focus on the localized errors by incorporating the classifier's prediction error into the noise schedule, which we describe in more detail in \cref{sec:classifier}. As shown in the first row of \cref{fig:motiv}, our refinement SB model (\ie{}, \Ours{}) corrects the distorted text in the initially generated image. 
% For example, we use bounding boxes to quickly annotate the noticeable artifacts in 5\% images generated by existing GANs (\cite{hrviton,gp-vton,sd-viton}). 
%While it may be possible to learn an artifact classifier by repurposing a GAN's fake/real discriminator, w

The contributions of our paper are:
\begin{itemize}   
   \setlength{\itemsep}{0pt}
  \setlength{\parskip}{0pt}
  \setlength{\parsep}{0pt}
    \item We introduce (human, synthetic garment) pairs as an augmentation in the training of virtual try-on task. The synthetic garment is obtained from our human-to-garment model, which can generate product-view garment images from human images.
    \item We introduce a spatially adaptive Schr\"odinger Bridge model (\Ours{}) to refine the outputs of a base virtual try-on model. Our formulation incorporates a \emph{spatially varying} diffusion noise schedule, with noise proportional to the degree of refinement we wish to perform locally. We find this to yield better results than the baseline Schr\"odinger Bridge framework.
    %It improves the low-quality region of an initially generated image based on a spatially-varying noise schedule that targets known artifacts.
    \item Extensive experiments on two datasets (VITON-HD \citep{hrviton} and DressCode-Upper \citep{dresscode}) show that \Ours{} enhances the quality of the images generated by prior work, and is preferred by the users in 59\% cases on average. 
\end{itemize}

\section{Related Work}
\paragraph{Training with Synthetic Data} The addition of synthetic data is often an effective means of improving downstream task performance when it is difficult to amass real data at the necessary scale. This has been demonstrated in the domains of image generation \citep{karras2020training,semantic_aug} and image editing \citep{brooks2023instructpix2pix,wasserman2024paint}. Careful applications can also be used to ameliorate dataset imbalance issues, as shown in \cite{dablain2022deepsmote}. Other works, such as \cite{alemohammad2024self}, use self-synthesized data to provide negative guidance for the diffusion model. 
Our incorporation of synthetic data in the virtual try-on task tackles a specific sub-problem in the broader image editing domain and is similar in spirit to \cite{brooks2023instructpix2pix,wasserman2024paint}. Specifically, we aim to synthesize paired training data that satisfies the stringent requirements of virtual try-on paired training data -- a canonical product-view garment image paired with an example of it being worn. Images of people in clothing are readily available, but it is difficult to obtain a product-view image of the exact clothing they are wearing.
To address this, our work tackles the human-to-garment problem, which aims to extract the clothing from a person's photo and project it to the canonical product view, making it roughly symmetric to the virtual try-on task. 
%\smallskip

\paragraph{Virtual Try-On} There has been a shift from earlier GAN-based framework \citep{clothflow,hrviton,sd-viton,gp-vton,keypoints-tryon} to diffusion-based methods \citep{stableviton,catvton} in the virtual try-on literature.
Diffusion models fit an SDE process mapping from the image distribution to the noise distribution, and tend to be easier to train than GAN-based approaches due to the simplicity of the L2 denoising loss \citep{ddim,diff_beat_gan,score_match}. At inference, the diffusion model denoises a random Gaussian noise distribution to a human-readable image via multiple sampling steps. \cite{tpd,catvton} propose parameter-efficient approaches that concatenate the human image and the garment images along the spatial dimension such that the self-attention layer in the denoising UNet can achieve texture transfer without extra parameters.  In \citep{stableviton}, the authors introduce additional cross-attention layers to learn the semantic correspondences between the garment and the human image. The methods in \citep{ning2024picture,unihuman,baldrati2023multimodal} align different embedding spaces in the attention module to achieve flexible clothing editing after try-on, such as style change or graphics insertion. In contrast to prior work that samples from random noise, we build upon recent advances in \Sch{} bridges, notably \cite{i2sb}, to directly sample from an initial image generated by prior try-on models. Our work is similar in spirit to \cite{cat-dm}, which initializes the noisy image with a GAN-generated image and small amounts of random noise. However, our work explores varying the local noise schedule based on the error level at a given location.
%we consider the \Sch{} bridge formulation to be a more principled approach to mapping from a pre-existing result. 

% \section{Method}
% Virtual try-on aims to resynthesize a given human image with a newly specified garment. We study the feasibility of improving current virtual try-on models by addressing both the data scarcity issue as well as through a refinement model designed to find-and-fix localized generation errors. We describe our approach to generating synthetic training data in \ref{sec:synthetic_data}, and then our EARSB refinement model in \ref{sec:earsb}.
%We wish to maintain the original pose of the human model while plausibly depicting how the specified garment would appear when worn on said model's body. 
%\section{Virtual Try-On Task Definition}

%One can first preprocess $x_0$ by masking out the target garment area using its segmentation map and also obtaining a pose representation for the masked area using DensePose \citep{guler2018densepose}. We denote the masked image as $\bar{x}_0$ and the pose representation as $P$ (see the left-most preprocessing step of Fig.  \ref{fig:fig2a}). Prior try-on work reconstructs $x_0$ using $\bar{x}_0$ conditioned on $C$ and $P$ \citep{cat-dm,catvton,idmvton}. 

\section{Augmented Training with Synthetic Data}
\label{sec:synthetic_data}
%The training data for a virtual try-on model comes in the form of human-garment image pairs $(x_0,C)$, where $x_0$ is the human image and $C$ is the canonical product-view image of the garment that the person is wearing. Given these pairs, one can mask out the target garment area from the ground-truth model's image, then train a try-on model to reconstruct $x_0$, conditioned on $C$.

\paragraph{Virtual Try-On Task Definition}
Let $(x_0,C)$ be the (human image, product-view of worn garment) pair in virtual try-on training.  We will refer to $(x_0, C)$ as paired data. Let $\bar{x}_0$ be a masked version of $x_0$ in which the worn garment corresponding to $C$ is masked out. We can then set up a learning task in which we aim to fit the following function:  $F(\bar{x}_0, C, \phi; \theta) \to x_0$, where $\phi$ corresponds to other conditionals such as pose representations from DensePose \citep{guler2018densepose}.

\begin{figure}
    \centering
        \begin{subfigure}[c]{0.5\textwidth}
        \centering
        \includegraphics[width=\textwidth]{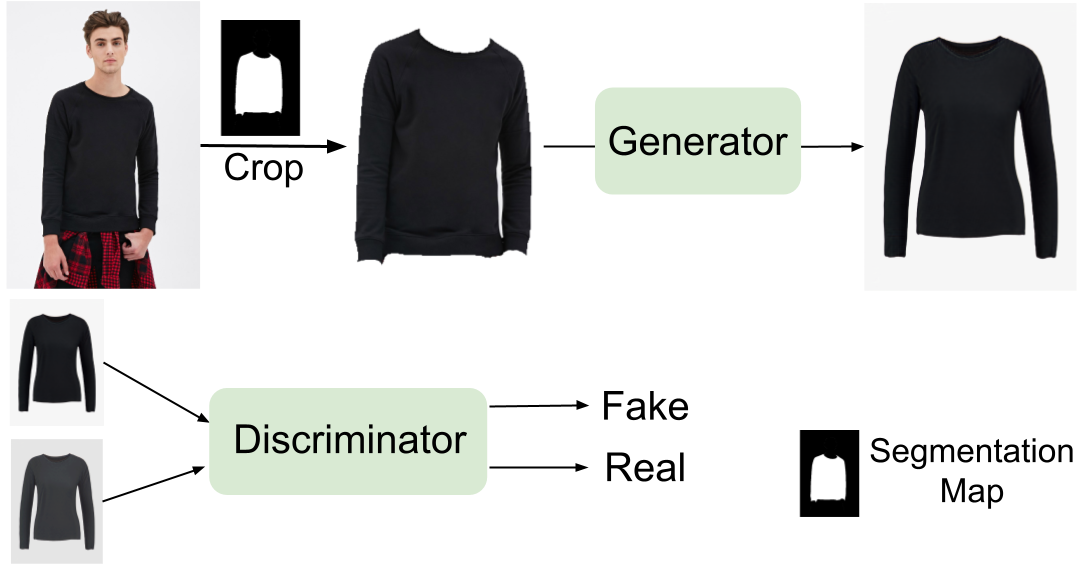}        
        \caption{Human-to-Garment model.}
        \label{fig:h2g}
    \end{subfigure}
    \hfill
    \begin{subfigure}[c]{0.5\textwidth}
        \centering
        \includegraphics[width=\textwidth]{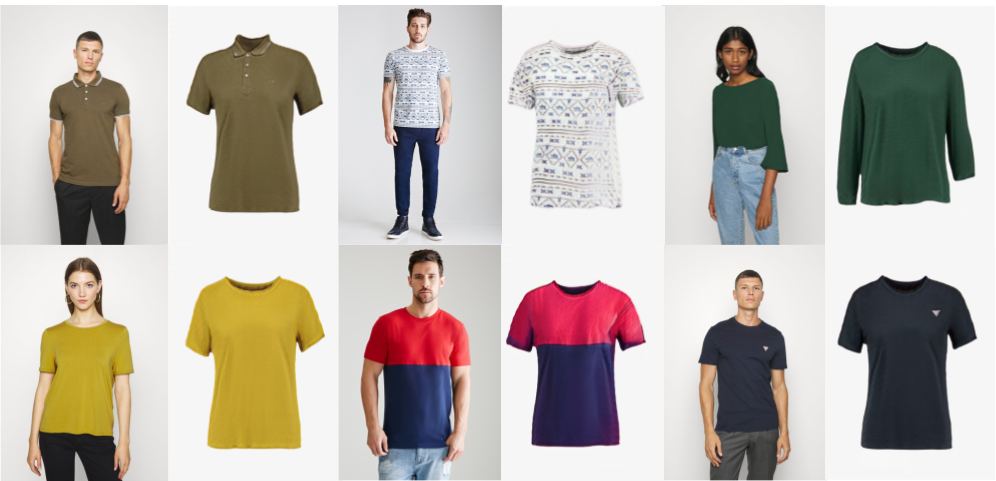}       
        \caption{Examples in H2G-UH and H2G-FH.}
        \label{fig:data_aug}
    \end{subfigure}
    \caption{(a) Our human-to-garment model, which is explained in \cref{sec:h2g}  (b) Examples of the constructed (human, synthetic garment) pairs in \cref{sec:h2g}.}
    \label{fig:aug_train}
    \vspace{-1em}
\end{figure}

Acquiring high-quality pairs $(x_0, C)$ at scale is challenging due to copyright and brand protection, but acquiring images of just humans ($x_0$) at scale is considerably more feasible \citep{deepfashion,sshq,upt,unihuman}. This observation motivates proposed human-to-garment process to extract a synthetic canonical view image $\hat{C}$ from $x_0$. We can then augment our virtual try-on training with $(x_0, \hat{C})$ pair, requiring only single human images. %These pairs can subsequently be utilized to augment the training data for our garment-on-human task (\ie{}, virtual try-on). 
In the following, \cref{sec:h2g} explains the architecture of our human-to-garment model, \cref{sec:gen_data} discusses how we use this model to construct the synthetic dataset, and \cref{sec:vtontraining} describes how the synthetic data is used to augment the virtual try-on training. 

\subsection{Human-to-Garment Model}
\label{sec:h2g}

While virtual try-on requires generating skin and deforming the product-view garment to accommodate diverse postures, the human-to-garment task simply aims to map the clothing item to its canonical view. To achieve this, we use existing paired (human, garment) data (\eg, VITON-HD \citep{hrviton}) to train our human-to-garment model. As illustrated in \cref{fig:h2g}, we first segment and extract the clothing on the person map and then feed the clothing item to a generator that synthesizes its canonical view. The generator is based on the UNet model proposed in \citep{clothflow}, which uses a flow-like mechanism for warping latent features in an optical-flow-like manner. The generator was trained using a combined L1 reconstruction and adversarial loss.

\subsection{Constructing Synthetic H2G-UH and H2G-FH}
\label{sec:gen_data}
%With the above human-to-garment model, we can obtain the product-view garments from single-human images \citep{deepfashion,sshq,upt} to create $(x_0, \hat{C}$ pairs. 
Synthetic images $\hat{C}$ produced from our models necessarily contain generation errors.
We use the following criteria to filter for high-quality synthetic data: a) The single human image $x_0$ has a clean background (low pixel variance in the non-human region); b) $x_0$ is frontal view (classified by its DensePose representation \citep{guler2018densepose}); c) the reconstruction error (LPIPS distance) is small when reconstructing the human image $x_0$ in a try-on model using the $(x_0, \hat{C})$ pair (\eg{}, \citep{hrviton,sd-viton}). Under these criteria, we select human images from DeepFashion2 \citep{deepfashion} and UPT \citep{upt}, eventually creating 12,730 synthetic pairs of upper-body human images (referred to as H2G-UH) and 8,939 pairs of full-body human images (referred to as H2G-FH). Examples of the synthetic pairs are shown in \cref{fig:data_aug}.

\subsection{Augmented Virtual Try-on Training}
\label{sec:vtontraining}
To further prevent distribution leakage of incorporating synthetic data, we explore two means of limiting the effect of the real-synthetic domain gap:
(a) two training stages involving pretraining the try-on model using synthetic pairs, and then finetuning on real pairs \citep{finetune};
(b) training simultaneously on real and synthetic data, but conditioning the try-on model on a real/synthetic flag, similar to \citep{jun2020distribution}. 
%The difference between these two is that in the first augmentation, the model only sees synthetic data at an early training stage, while in the second augmentation, the model takes both real and synthetic pairs throughout the training. 
We found empirically that the second augmentation performs slightly better than the first (See \cref{sec:quan_results}). 

\begin{figure*}[!t]
    \centering    
    \includegraphics[width=\textwidth]{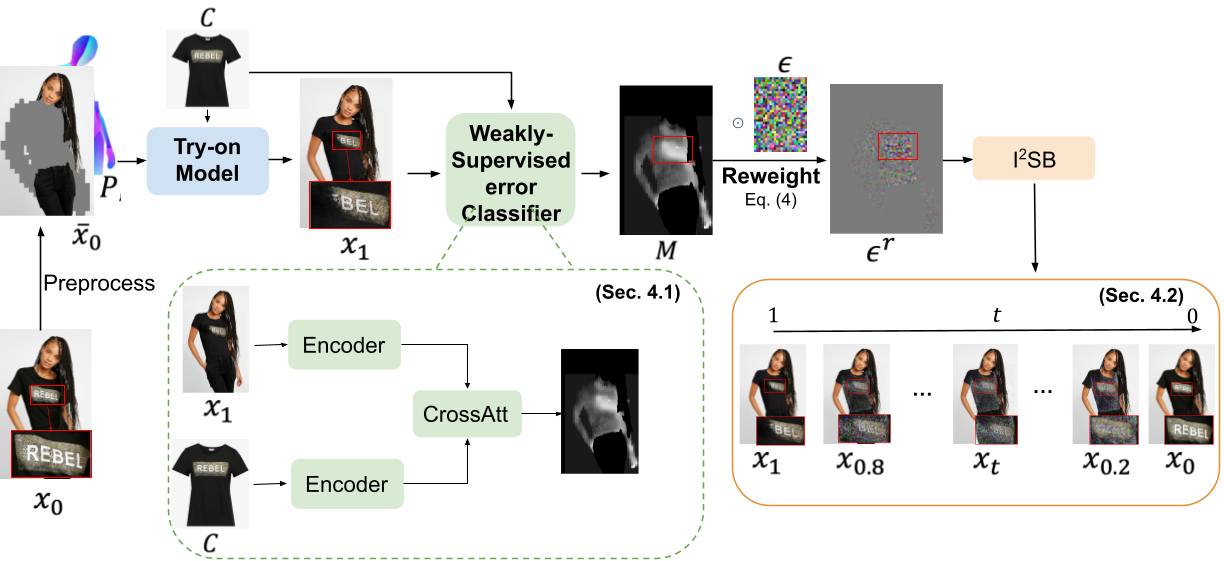}    
    \caption{The diffusion process in our refinement-based \Ours{}. We first preprocess the input image, then use a base try-on model that takes the masked human image $\bar{x}_0$, its pose representation $P$, and its garment $C$ as input to generate an initial human image $x_1$. $x_1$ is fed to our weakly-supervised classifier (WSC) to obtain the error map $M$ (see \cref{sec:classifier}). This map reweights the noise distribution $\epsilon$ to $\epsilon^r$ in I$^2$SB diffusion and refines $x_1$ that has generation errors to the ground truth image $x_0$ (see \cref{sec:pipeline}).}
     \label{fig:fig2a}
     \vspace{-1em}
\end{figure*}

\section{Error-Aware Refinement \Sch{} Bridge}
\label{sec:earsb}
Apart from the synthetic data augmentation from \cref{sec:synthetic_data}, our second approach to enhancing existing try-on methods is a refinement pipeline with two steps. %: first identifying the errors in the generated try-on image, referred to as $x_1$, and then fitting a targeted model that learns to refine the erroneous locations in $x_1$. 
First, given some base model $F_{base}(\bar{x}_0, C, \phi) \to x_1$ where $F_{base}()$ can be any pretrained GAN or diffusion-based approach to virtual try-on, $x_1$ closely approximates the true real human image $x_0$ with some generation artifacts. To automatically identify the artifacts in $x_1$, we construct a weakly-supervised error classifier $\text{WSC}(x_1,C) \to M$ as in \cref{fig:fig2a}, where $M$ is a confidence map predicting a heatmap for likely generation errors. Then, our second step performs the final refinement by fitting a \Sch{} bridge based on I$^2$SB \cite{i2sb} via the following mapping: $F_{EARSB}(x_1, C, M, \phi; \theta) \to x_0$.
 
%In the scope of this paper, we focus on the virtual try-on task.  

%I$^2$SB constructs a \Sch{} Bridge (SB) that allows us to map from data to data in accordance with our goals.
%Let $x_1$ be a virtual try-on rendering from a try-on model \citep{sd-viton,stableviton}, and $x_0$ is the ground-truth target image. We wish to construct a refinement model that maps from $x_1$ to $x_0$. (See bottom right of \cref{fig:fig2a}). 
%(possibly trained with augmented synthetic data as described in \ref{sec:vtontraining})

The approach is weakly inspired by boosting methods in that we wish to fit a \emph{targeted} refinement model that is trained specifically on the generation errors of an existing model. The refinement goal applies to the general setting where we want to refine a flawed image output, though we focus on virtual try-on for this work.  Thus, as illustrated in \cref{fig:fig2a}, the training of \Ours{} includes three steps:
\begin{enumerate}[leftmargin=2\parindent]
    \item  Pre-process the images in the training set and feed them to existing try-on models to get the initial images $x_1$.
    \item Obtain the error maps $M$ on the initial images $x_1$ using our WSC (\cref{sec:classifier}).
    \item Use $M$ to adjust the noise schedule in I$^2$SB \cite{i2sb} and train the noise prediction model in \Ours{} following \cref{eq:loss} (\cref{sec:pipeline}).
\end{enumerate}

As the first step simply requires running an off-the-shelf virtual try-on model to obtain an initial image, our discussion in the next section will begin by describing the second step- obtaining the error map.
%In the following, we start with explaining step 2 as step 1 is simply running prior try-on models.

%This is ideal in the case of an area correctly rendered in $x_1$.
%It would have to learn this implicitly from its training.
%At the same time, an area assigned a low noise volume (low confidence of error) indicates high quality and will have reduced capacity for altering the original pixel values. 

\subsection{Obtaining the Error Map}
\label{sec:classifier}
We start by obtaining the error map $M$ that highlights the corrupted or incorrect area of the initial image $x_1$ using our proposed Weakly-Supervised error Classifier (WSC).% to predict this error map.
%We found that an effective weakly-supervised classifier (WSC) can be constructed to predict this error map with just a few hours of manual labeling of generation errors, and was more straightforward to construct than adapting a GAN's discriminator.

\paragraph{Classifier Architecture}
As shown in the green dotted box of \cref{fig:fig2a}, our WSC has two encoders to match the image features of $x_1$ and $C$ with cross attention to predict a sigmoid-activated error map.  

\paragraph{Training Data Annotation}
%Ideally, the classifier should predict 1 for GAN-generated low-quality regions and 0 for error-free regions in $M$. However, we do not have full annotations for where all the generated artifacts are located. Further, automated metrics such as localized reconstruction errors would produce too many false positives, as they do not account for the full distribution of valid renderings. 
In practice, it is labor-intensive to fully annotate all the initial images for where the generated artifacts are located. 
To mitigate this issue, we used a few hours to hand-label a small portion of the initial try-on images in the training set at the patch level, using bounding boxes for poorly generated regions. 

\paragraph{Weakly-Supervised Training}
%Our WSC is trained with patch-level supervision for labeled images, and with image-level supervision (generated vs real) for unlabeled images.
Let $x_0,x_1^u,x_1^l$ be the real human image, the unlabeled initial image, and the labeled initial image with bounding boxes annotating artifacts. Our WSC loss terms are defined as:
\begin{equation}
\begin{aligned}
    \mathcal{L}_{img}=&-\log{\big{(}{\text{WSC}}(x_1^u,C)_{\text{max}}\big{)}}+\log{\big{(}1-\text{WSC}(x_0,C)_{\text{max}}\big{)}} \\
    \mathcal{L}_{pat}=&-\log{\big{(}\text{WSC}(x_1^l,C)\odot B_{box}\big{)}} \\
    &-\log{\big{(}1-{\text{WSC}}(x_1^l,C)\odot (1-B_{box}) \big{)}}
\end{aligned}
\end{equation}
where $\mathcal{L}_{img}$ is the image-level loss and $\mathcal{L}_{pat}$ is the patch-level loss. In $\mathcal{L}_{img}$, $\text{WSC}(\cdot)$ is the output error map and $\text{WSC}(\cdot)_{\text{max}}$ denotes the spatially max-pooled score in the error map. In $\mathcal{L}_{pat}$, $B_{box}$ is the spatial binary mask for the annotated regions, thereby maximizing and minimizing the scores for regions within and outside of the annotated boxes respectively. Our final loss is: $\mathcal{L}_{\textrm{WSC}} ={\mathcal{L}_{ins}+\mathcal{L}_{pat}} $.
%This can also be interpreted as a multi-instance learning problem where the training model only knows that at least one region within the image is of low quality. The patch-level loss $\mathcal{L}_{pat}$ further introduces positive and negative pixel samples and is only available for fully annotated samples. 
%In practice, our classifier operates and averages results over 3 different feature map resolutions, with additional upsampling for lower resolutions to ensure the same output dimensions.

The trained WSC will predict an error map $M$ for the initial image $x_1$, which is then used to adjust the noise schedule in the diffusion process described in the next section.

\subsection{Error-Map-Reweighted SB Formulation}
\label{sec:pipeline}
%Let $x_1$ be the reconstructed image from a base try-on model. We wish to identify the artifacts in $x_1$ and construct a refinement model that maps from $x_1$ to $x_0$. (See bottom right of \cref{fig:fig2a}).

To achieve the refinement goal, our diffusion process extends \Sch{} bridges as formulated in I$^2$SB \citep{i2sb}, where we incrementally add noise to the initial image $x_1$, and then remove the noise to approximate the refined image $x_0$.
%One interpretation of paired \Sch{} is that of stochastic interpolation \cite{albergo2023stochastic}.
%A na\"ively trained SB model is provided training pairs ($x_0$, $x_1$) to map between, but 
However, without additional information, a na\"ively trained I$^2$SB model must implicitly learn what to refine and what to retain. Our formulation aims to explicitly incorporate prior knowledge of localized generation errors via the error map $M$ into the \Sch{} process by using $M$ to locally scale the noise schedule for the \Sch{} process.

Our choice of locally scaling the noise schedule is based on several observations. We want the model to directly copy pixels over to $x_0$ for correctly generated regions in $x_1$ -- it would be nice to avoid training the model to add and remove noise from these regions. In contrast, erroneous regions in the initial try-on images, especially more noticeable ones, include generation errors that share little to no structural similarity to the target. These errors include examples such as deformed limbs, and distorted textures/fabrics, which may need more added noise to prevent the model from conditioning too strongly on the original pattern.

%Empirically, the magnitude of changes between $x_0$ to $x_1$ should be proportional to the noise level, and not all regions in our setup will require refinement. Based on this, we propose a 
%more \emph{targeted} approach where we first identify the errors in $x_1$ with an error map $M$, and then make the model directly \emph{learn} the refinement based on the error map with \emph{spatially adaptive} noise schedule. 

As such, we construct our refinement model $F_{EARSB}(x_1, C, M, P; \theta) \to x_0$, where the model is conditioned on the canonical view garment $C$, the error map $M$, and the pose representation $P$. \cref{fig:fig2a} shows our Weakly-Supervised Classifier (WSC) first locates the errors in the error map $M$, then $M$ \emph{reweights} the noise schedule of the I$^2$SB stochastic process to assign a higher volume of noise to the low-quality region ``rebel" so the model can focus on refining it.

 \paragraph{Error-Map-Reweighted Diffusion Process} 
 Following I$^2$SB \citep{i2sb}, our diffusion \Sch{} bridge maps from the initial image $x_1$ to the ground truth image $x_0$. It fits to the following stochastic process:
 \begin{equation}
\begin{gathered}
    x_t=\mu_t(x_0,x_1)+\sqrt{\Sigma_t} \cdot \epsilon \\   
    \mu_t=\frac{\bar{\sigma}_t^2}{\bar{\sigma}_t^2+\sigma_t^2}x_0
    +\frac{\sigma_t^2}{\bar{\sigma}_t^2+\sigma_t^2}x_1 , \text{ }\Sigma_t=\frac{{\bar{\sigma}_t^2}\sigma_t^2}{\bar{\sigma}_t^2+\sigma_t^2} \cdot I ,
\end{gathered}
\label{eq1}
\end{equation}
where $\sigma_t^2=\int_{0}^{t}{\beta_\tau} d \tau$, $\bar{\sigma}_t^2=\int_{t}^{1}{\beta_\tau} d \tau $ and
${\beta_\tau}$ is a symmetrical noise schedule. $\epsilon \sim \mathcal{N}(0,I)$ is random Gaussian noise. The above equation stochastically adds noise and then removes it between $x_1$ and $x_0$. %Empirically, the magnitude of changes between $x_1$ to $x_0$ should be proportional to the noise level, and not all regions in our setup will require refinement. Inspired by this intuition, we propose the error-map-reweighted noise schedule in the following. 

% \begin{figure}
% \centering
% \includegraphics[width=0.8\linewidth]{figures/fig2b.png}        \vspace{-1em}
%     \caption{\Ours{}'s denoising UNet (see \cref{sec:pipeline}). Inputs are the concatenation of $M$, $P$ and $x_1$; outputs are the predicted reweighted noise $\epsilon^r_\theta$.}
%     \label{fig:fig2b}
%     \vspace{-1em}
% \end{figure}

%The major difference between the propsoed \Ours{} and I$^2$SB is our error-map reweighted diffusion process.
%Noise schedules in I$^2$SB typically vary only with the time variable $t$. 
We extend I$^2$SB such that the noise schedule can vary spatially based on the error map $M$ (obtained from WSC in \cref{sec:classifier}). Good regions will be assigned less noise (\ie{}, smaller variance) in the diffusion process, while poor quality regions will be assigned more:
%. For example, mono-color region tends to have fewer artifacts than complex graphic parts on the clothing. 
\begin{gather}
    x_t=\mu_t(x_0,x_1)+\sqrt{\Sigma_t} \cdot \epsilon^r, \\
    \epsilon^r=M \cdot \epsilon, M=\text{WSC}(x_1, C)
\end{gather}
where $\mu_t$ is the same as  \cref{eq1} and $\epsilon^r$ is the adaptive noise. %$M$ highlights low-quality regions in $x_1$ and assigns noise with higher variance to them, thus steering the model to learn rich textures with high fidelity.

\paragraph{Sampling Process} 
%As shown in \cref{fig:fig2b}
%encodes the concatenation of the error map $M$, the pose representation $P$, and the noisy image $x_t$, A separate encoder is used to learn the feature representation of the given garment 
The initial image $x_1$ is iteratively refined to $x_0$ via a denoising/sampling process, where a model predicts the noise distribution at each time step. In contrast to prior soft-attention-based UNets \citep{cat-dm,tpd,stableviton,ladiviton}, our denoising model uses cloth-flow-learning UNet for more precise garment deformation \citep{clothflow}. It accepts the garment $C$, the error map $M$, the pose representation $P$, and the noisy image $x_t$ as inputs and predicts the error-adapted noise $\epsilon^r_\theta(\cdot;t)$, where $(\cdot;t)$ omits the inputs $M,P,x_t,C$. See Supp. B for the detailed model architecture. With the predicted noise $\epsilon^r_\theta(\cdot;t)$, we define our sampling process:
\begin{gather}
\hat{x}_0=x_t-\sqrt{\Sigma_t} \cdot \epsilon^r_\theta(M,P,x_t,C;t) 
    \label{eq:bkwd} \\
    x_{t-{\Delta}t}=\hat{\mu}_{t-{\Delta}t}(\hat{x}_0,x_t) + M \cdot \sqrt{\hat{\Sigma}_t} \cdot \epsilon \\
    \hat{\mu}_{t-{\Delta}t}=\frac{\sigma_{t-{\Delta}t}^2}{\sigma_t^2} \hat{x}_0 + \frac{\sigma_t^2 - \sigma_{t-{\Delta}t}^2}{\sigma_t^2} x_t  \\
    \hat{\Sigma}_t = \frac{\sigma_{t-{\Delta}t}^2(\sigma_t^2 - \sigma_{t-{\Delta}t}^2)}{\sigma_t^2}
\end{gather}
where ${\Delta}t>0$ and it is the sampling interval. Starting from $t=1$, the process iteratively refines the initial human image $x_1$ based on the error map $M$. When $M$ is all ones in \cref{eq:bkwd}, our model reverts to the I$^2$SB formulation. When $M$ is all zeros (\ie{}, no error), $x_1$ is believed to be perfect $x_1$ does not need to be refined in the sampling process.
%so our UNet should predict zero noise in $\epsilon^r_\theta$ and 

The training objective of our model is the mean squared error between the predicted noise $\epsilon^r_\theta$ and the reweighted Gaussian noise $\epsilon^r$
\begin{equation}
    \mathcal{L}_{\mathrm{EARSB}}=\mathbb{E}_{t \sim U(0,1)}|| \epsilon^r_\theta(M,P,x_t,C;t)- \epsilon^r||^2
    \label{eq:loss}
\end{equation}

\subsubsection{Further Improvements via Classifier Guidance and Expert Denoisers}
%where the classifier often refers to an object category classifier 
%Classifier guidance has been demonstrated to improve the image fidelity in diffusion models \citep{diff_beat_gan}. In our formulation of WSC, the classifier guidance 
Whereas prior work used an object category classifier to guide the sampling process \citep{diff_beat_gan}, our WSC guidance gives a direction toward the real data distribution. \citet{chung2023diffusion} shows  we can estimate the guidance score $\nabla_{x_t}\log{p(y|x_t)}$ using the denoised clean image $\hat{x}_0$:  $\nabla_{x_t}\log{p(y|x_t)} \simeq \nabla_{x_t}\log{p(y|\hat{x}_0)} $, where $y$ is the fake/real label. Since the label for real data is 0 in WSC, the classifier guidance is:
\begin{equation}
    \hat{\mu}_{t-{\Delta}t} \leftarrow \hat{\mu}_{t-{\Delta}t}+ M \cdot \hat{\Sigma}_t \cdot \nabla_{x_t}\log{p(\mathbf{0}|\hat{x}_0)}
    \label{eq:cg}
\end{equation}
where $p(\mathbf{0}|\hat{x}_0)=1-\text{WSC}(\hat{x}_0,C)$. %We found scaling the guidance score to be larger and clamping its value (\eg{}, $[-0.1, 0.1]$) helps strengthen the guidance without image quality loss.

\begin{table*}[!t]
    \centering
    %\small
    \setlength{\tabcolsep}{1pt}
    \begin{tabular}{@{}rlcccccccccccc@{}}
        \toprule     
        &&\multicolumn{6}{c}{VITON-HD} 
        &\multicolumn{6}{c}{DressCode-Upper} \cr
         \cmidrule(r){3-8} \cmidrule(l){9-14} 
        && \multicolumn{2}{c}{Unpaired} 
        & \multicolumn{4}{c}{Paired}
        & \multicolumn{2}{c}{Unpaired} 
        & \multicolumn{4}{c}{Paired}\cr
        \cmidrule(r){3-4} \cmidrule(lr){5-8}
        \cmidrule(lr){9-10} \cmidrule(l){11-14} 
        && FID$\downarrow$  & KID$\downarrow$ & FID$\downarrow$  & KID$\downarrow$ & SSIM$\uparrow$ & LPIPS$\downarrow$ 
        & FID$\downarrow$  & KID$\downarrow$ & FID$\downarrow$  & KID$\downarrow$ & SSIM$\uparrow$ & LPIPS$\downarrow$ 
        \cr
        \midrule
        \textbf{(a)} & \textbf{GAN-Based}\cr
        & HR-VTON \citep{hrviton} &10.75 &0.28 &8.46 &0.26 &0.901 &0.075 &15.26 &0.39 &11.76 &0.32 &0.947 &0.046\cr
        &SD-VTON \citep{sd-viton} &9.05 &0.12 &6.47 &0.09 &0.907 &0.070 &14.73 &0.32 &10.99 &0.24 &0.947 &0.042\cr
        &GP-VTON \citep{gp-vton} &8.61 &0.86 &5.53 &0.07 &0.913 &0.064 &26.19 &1.71 &23.66 &1.59 &0.816 &0.262\cr
        &\Ours{} (ours) &8.42 &0.07 &5.25 &0.05 &0.918 &0.059 &10.89 &0.13 &7.15 &0.13 &0.961 &0.028 \cr
        %\Ours{} pre. H2G-UH/FH &8.35 &0.07 &5.18 &0.05 &0.918 &0.059 &10.79 &0.12 &7.10 &0.12 &0.958 &0.028\cr
        %\Ours{} + plain H2G-UH/FH  &9.64 &0.15 &6.52 &0.11 &0.902 &0.073 &11.52 &0.21 &8.56 &0.15 &0.950 &0.038\cr
        &\Ours{} +H2G-UH/FH (ours) & \textbf{8.26} &\textbf{0.06} &\textbf{5.14} &\textbf{0.04} &\textbf{0.919} &\textbf{0.058}  &\textbf{10.70} &\textbf{0.11} &\textbf{7.05} &\textbf{0.11} &\textbf{0.965} &\textbf{0.026} \cr
        \midrule
        \textbf{(b)} & \textbf{SD-Based}\cr
        &LaDI-VTON \citep{ladiviton} &8.95 &0.12 &6.05 &0.08 &0.902 &0.071 &14.88
&0.39 &11.61 &0.32 &0.939 &0.057 \cr
        &CatVTON \citep{catvton} &8.87 &0.08 &5.49 &0.07 &0.915 &0.059 &11.91 &0.21  &7.66 &0.10 &0.950 &0.038\cr
        &CAT-DM \citep{cat-dm} &8.55 &0.10 &5.98 &0.07 &0.908 &0.067
           &12.91 &0.29 &8.58 &0.16 &0.948 &0.038 \cr
        &IDM-VTON \citep{idmvton} &8.59 &0.11 &5.51 &0.09 &0.902 &0.061 &11.09 &0.16 &6.79 &0.12 &0.956 &0.026 \cr % results will come out tonight
        &TPD \citep{tpd} &8.23 &\textbf{0.06} &\textbf{4.86} &0.04 &0.917 &0.057
            &\multicolumn{6}{c}{-} \cr
        &StableVITON \citep{stableviton} &8.20 &0.07 &5.16 &0.05 &0.917 &0.057 &\multicolumn{6}{c}{-}\cr 
        &\Ours{}(SD) +H2G-UH/FH (ours) &\textbf{8.04} &\textbf{0.06} &4.90 &\textbf{0.03} &\textbf{0.925} &\textbf{0.053} 
        &\textbf{10.41} &\textbf{0.09} &\textbf{6.76} &\textbf{0.08} &\textbf{0.968} &\textbf{0.023} \cr
        %\midrule
        \bottomrule
    \end{tabular}
    \vspace{-.5em}
    \caption{Virtual try-on results on VITON-HD \citep{hrviton} and DressCode-Upper \citep{dresscode} for \textbf{(a)} GAN-based and \textbf{(b)} diffusion-based models using 25 sampling steps. KID is multiplied by 100. We find our \Ours{} approach outperforms prior work on average. See \cref{sec:quan_results} for discussion.}
    \label{tab:metrics}
    \vspace{-1em}
\end{table*}

Following~\citep{balaji2022eDiff-I}, a trained \Ours{} model is split into two models, each having an expert denoiser that is fine-tuned on denoising ranges $t\in [0, 0.5]$ and $t \in [0.5, 1]$, respectively. %We use this method in all  variants in the next section.
%as we find it improves results with no trade-offs (other than model size).

\section{Experiments}
\paragraph{Datasets}
We use VITON-HD \citep{hrviton}, DressCode-Upper \citep{dresscode}, and our synthetic H2G-UH and H2G-FH for training. They include 11,647, 13,564, 12,730, 8,939 training images, respectively. For synthetic data augmentation, we combine VITON-HD with our H2G-UH since they both include mostly upper-body human images. DressCode-Upper is combined with H2G-FH as both consist of full-body human photos.
For evaluation, VITON-HD contains 2,032 (human, garment) test pairs and DressCode-Upper has 1,800 test pairs and include both paired and unpaired settings. In the paired setting, the input garment image and the garment in the human image are the same item. Conversely, the unpaired setting uses a different garment image.
%DressCode-Upper takes the upper clothing category from the original DressCode dataset. 

\paragraph{Metrics}
We use Structural Similarity Index Measure (SSIM) \citep{ssim}, Frechet Inception Distance (FID) \citep{fid}, Kernel Inception Distance (KID) \citep{kid}, and Learned Perceptual Image Patch Similarity (LPIPS) \citep{lpips} to evaluate image quality. All the compared methods use the same image size $512\times 512$ and padding when computing the above metrics.

\paragraph{Experimental setup} We compare \Ours{} with GAN-based methods HR-VTON \citep{hrviton}, SD-VTON~\citep{sd-viton} and GP-VTON~\citep{gp-vton}, as well as Stable Diffusion (SD) based methods including CAT-DM \citep{cat-dm}, StableVITON \citep{stableviton}, TPD \citep{tpd}, IDM-VTON \citep{idmvton} and CatVTON \citep{catvton}. Unless otherwise specified, all diffusion models use 25 sampling steps. %The results under different sampling steps on VITON-HD are presented in \cref{fig:ss}. 

We report results of multiple variants of our approach. \Ours{} uses GAN-based GP-VTON \citep{gp-vton} to generate the initial image that was trained without synthetic data augmentation.  \Ours{}+H2G-UH/FH trains with either H2G-UH or H2G-FH. We add the upper-body synthetic subset H2G-UH for the upper-body-human dataset VITON-HD, and the full-body synthetic H2G-FH when on DressCode-Upper. \Ours{}(SD)+H2G-UH/FH uses the diffusion-based CatVTON \citep{catvton} to generate the initial image.

\subsection{Results}
\label{sec:quan_results}

\cref{tab:metrics} compares our approach with those from prior work. We find our full model variants \Ours{}+H2G-UH/FH and \Ours{}(SD)+H2G-UH/FH boosts performance over the GAN and SD-based methods, respectively. The last two rows of \cref{tab:metrics}(a) show that incorporating our synthetic training pairs provide a consistent boost in performance on both datasets. Comparing the last rows of \cref{tab:metrics}(a) and \cref{tab:metrics}(b) we observe that using the diffusion model to generate the initial image gives better performance in \Ours{}(SD)+H2G-UH/FH, but is more costly. %Overall, our refinement model improves the metrics of existing methods on both datasets, with further improvements from incorporating synthetic training pairs.

% Note that all the compared SD-based methods in \cref{tab:metrics} use pretrained SD model weights, which are obtained by training on millions of (image, text) pairs, while our \Ours{}+H2G-UH/FH was trained from scratch on try-on datasets with only around 24k (human, garment) pairs. Still, \Ours{}+H2G-UH/FH achieves comparable performance on VITON-HD and even outperforms SD baselines on DressCode-Upper. We also experimented with adding our synthetic data into the training of SD-based methods in \cref{sec:abl} and observed similar improvements. %(see \cref{tab:abl}), demonstrating that synthetic data augmentation can be complimentary to the large-scale pretraining used by SD-based models.

\begin{table}[!t]
\centering
    \setlength{\tabcolsep}{1.3pt}
    %\scalebox{0.9}{
    \begin{tabular}{@{}lcccc@{}}
    \toprule
    Methods &GP-VTON &\Ours{} &StableVITON &\Ours{} \cr \cmidrule(r){2-3} \cmidrule(l){4-5} 
     Consistency &42\% &58\% &38\% &62\%\cr 
     Fidelity  &39\% &61\% &45\% &55\%\cr
    \bottomrule
    \end{tabular}
    %}
\vspace{-.5em}
\caption{User studies on VITON-HD. Our \Ours{} method is preferred in an average of 59\% cases.}
\label{tab:user}
\vspace{-2em}
\end{table}

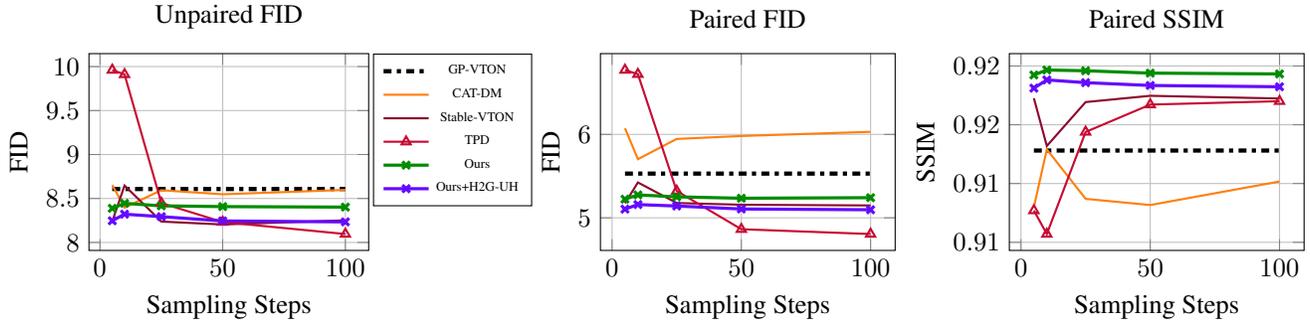
\begin{figure*}[!t]
    
    \begin{subfigure}[t]{0.4\textwidth}
    \definecolor{forestgreen}{RGB}{34, 139, 34}
\definecolor{color1}{RGB}{0, 0, 0} % blue
\definecolor{color2}{RGB}{255, 127, 14} % orange
\definecolor{color3}{RGB}{140, 10, 44}  % green
\definecolor{color6}{RGB}{100, 10, 255}  % light blue
\definecolor{color5}{RGB}{0, 140, 0} % purple
\definecolor{color4}{RGB}{200, 10, 50}  % brown

\begin{tikzpicture}
    \begin{axis}[
        title={Unpaired FID},
        xlabel={Sampling Steps},
        ylabel={FID},
        legend style={at={(1.3,1)}, anchor=north, font=\tiny},
        grid=major,
        xlabel near ticks,
        ylabel near ticks,
        width=5.3cm, height=4.2cm
    ]
        % Plot for the first CSV file
        \addplot[
            color=color1,
            ultra thick,
            dashdotted
        ] 
        table [col sep=comma, x index=0, y index=1, header=true] {figures/data/unpaired_steps_fid.csv};
        \addlegendentry{GP-VTON}

        % Plot for the second CSV file
        \addplot[
            color=color2,
            mark=dashed,thick
        ] 
        table [col sep=comma, x index=0, y index=2, header=true] {figures/data/unpaired_steps_fid.csv};
        \addlegendentry{CAT-DM}

        % Plot for the third CSV file
        \addplot[
            color=color3,
            mark=dashed,thick
        ] 
        table [col sep=comma, x index=0, y index=3, header=true] {figures/data/unpaired_steps_fid.csv};
        \addlegendentry{Stable-VTON}
         \addplot[
            color=color4,
            mark=triangle,thick
        ] 
        table [col sep=comma, x index=0, y index=4
        , header=true] {figures/data/unpaired_steps_fid.csv};
        \addlegendentry{TPD}
         \addplot[
            color=color5,
            mark=x,
            very thick
        ] 
        table [col sep=comma, x index=0, y index=5
        , header=true] {figures/data/unpaired_steps_fid.csv};
        \addlegendentry{Ours}
         \addplot[
            color=color6,
            mark=x,
            very thick
        ] 
        table [col sep=comma, x index=0, y index=6
        , header=true] {figures/data/unpaired_steps_fid.csv};
        \addlegendentry{Ours+H2G-UH}
    \end{axis}
\end{tikzpicture}
    \end{subfigure}
    \begin{subfigure}[t]{0.28\textwidth}
        \definecolor{forestgreen}{RGB}{34, 139, 34}
\definecolor{color1}{RGB}{0, 0, 0} % blue
\definecolor{color2}{RGB}{255, 127, 14} % orange
\definecolor{color3}{RGB}{140, 10, 44}  % green
\definecolor{color6}{RGB}{100, 10, 255}  % light blue
\definecolor{color5}{RGB}{0, 140, 0} % purple
\definecolor{color4}{RGB}{200, 10, 50}  % brown

\begin{tikzpicture}
    \begin{axis}[
        title={Paired FID},
        xlabel={Sampling Steps},
        ylabel={FID},
        grid=major,
        xlabel near ticks,
        ylabel near ticks,
        width=5.5cm, height=4.2cm
    ]
        % Plot for the first CSV file
        \addplot[
            color=color1,
            ultra thick,
            dashdotted
        ] 
        table [col sep=comma, x index=0, y index=1, header=true] {figures/data/paired_steps_fid.csv};
        %\addlegendentry{GP-VTON}

        % Plot for the second CSV file
        \addplot[
            color=color2,
            mark=dashed,thick
        ] 
        table [col sep=comma, x index=0, y index=2, header=true] {figures/data/paired_steps_fid.csv};
        %\addlegendentry{CAT-DM}

        % Plot for the third CSV file
        \addplot[
            color=color3,
            mark=dashed,thick
        ] 
        table [col sep=comma, x index=0, y index=3, header=true] {figures/data/paired_steps_fid.csv};
        %\addlegendentry{Stable-VTON}
         \addplot[
            color=color4,
            mark=triangle,thick
        ] 
        table [col sep=comma, x index=0, y index=4
        , header=true] {figures/data/paired_steps_fid.csv};
        %\addlegendentry{TPD}
         \addplot[
            color=color5,
            mark=x,
            very thick
        ] 
        table [col sep=comma, x index=0, y index=5
        , header=true] {figures/data/paired_steps_fid.csv};
        %\addlegendentry{Ours}
         \addplot[
            color=color6,
            mark=x,
            very thick
        ] 
        table [col sep=comma, x index=0, y index=6
        , header=true] {figures/data/paired_steps_fid.csv};
        %\addlegendentry{Ours+H2G-UH}
    \end{axis}
\end{tikzpicture}
    \end{subfigure}
    \begin{subfigure}[t]{0.28\textwidth}
        \definecolor{forestgreen}{RGB}{34, 139, 34}
\definecolor{color1}{RGB}{0, 0, 0} 
\definecolor{color2}{RGB}{255, 127, 14} 
\definecolor{color3}{RGB}{140, 10, 44}  
\definecolor{color6}{RGB}{100, 10, 255} 
\definecolor{color5}{RGB}{0, 140, 0} 
\definecolor{color4}{RGB}{200, 10, 50} 

\begin{tikzpicture}
    \begin{axis}[
        title={Paired SSIM},
        xlabel={Sampling Steps},
        ylabel={SSIM},
        grid=major,
        xlabel near ticks,
        ylabel near ticks,
        width=5.5cm, height=4.2cm
    ]
        % Plot for the first CSV file
        \addplot[
            color=color1,
            ultra thick,
            dashdotted
        ] 
        table [col sep=comma, x index=0, y index=1, header=true] {figures/data/paired_steps_ssim.csv};
        %\addlegendentry{GP-VTON}

        % Plot for the second CSV file
        \addplot[
            color=color2,
            mark=dashed,thick
        ] 
        table [col sep=comma, x index=0, y index=2, header=true] {figures/data/paired_steps_ssim.csv};
        %\addlegendentry{CAT-DM}

        % Plot for the third CSV file
        \addplot[
            color=color3,
            mark=dashed,thick
        ] 
        table [col sep=comma, x index=0, y index=3, header=true] {figures/data/paired_steps_ssim.csv};
        %\addlegendentry{Stable-VTON}
         \addplot[
            color=color4,
            mark=triangle,thick
        ] 
        table [col sep=comma, x index=0, y index=4
        , header=true] {figures/data/paired_steps_ssim.csv};
        %\addlegendentry{TPD}
         \addplot[
            color=color5,
            mark=x,
            very thick
        ] 
        table [col sep=comma, x index=0, y index=5
        , header=true] {figures/data/paired_steps_ssim.csv};
        %\addlegendentry{Ours}
         \addplot[
            color=color6,
            mark=x,
            very thick
        ] 
        table [col sep=comma, x index=0, y index=6
        , header=true] {figures/data/paired_steps_ssim.csv};
        %\addlegendentry{Ours+H2G-UH}
    \end{axis}
\end{tikzpicture}
    \end{subfigure}
    \vspace{-2em}
    \caption{Results on VITON-HD at 5, 10, 25, 50, and 100 sampling steps. Our method consistently improves our baseline starting model GP-VTON (black, dotted line), making it competitive with StableVITON (especially at under 50 sampling steps). Legend is shared for all.}
    \label{fig:ss}
    \vspace{-.7em}
\end{figure*}

\begin{figure*}[!t]
     \centering
    \includegraphics[width=\textwidth]{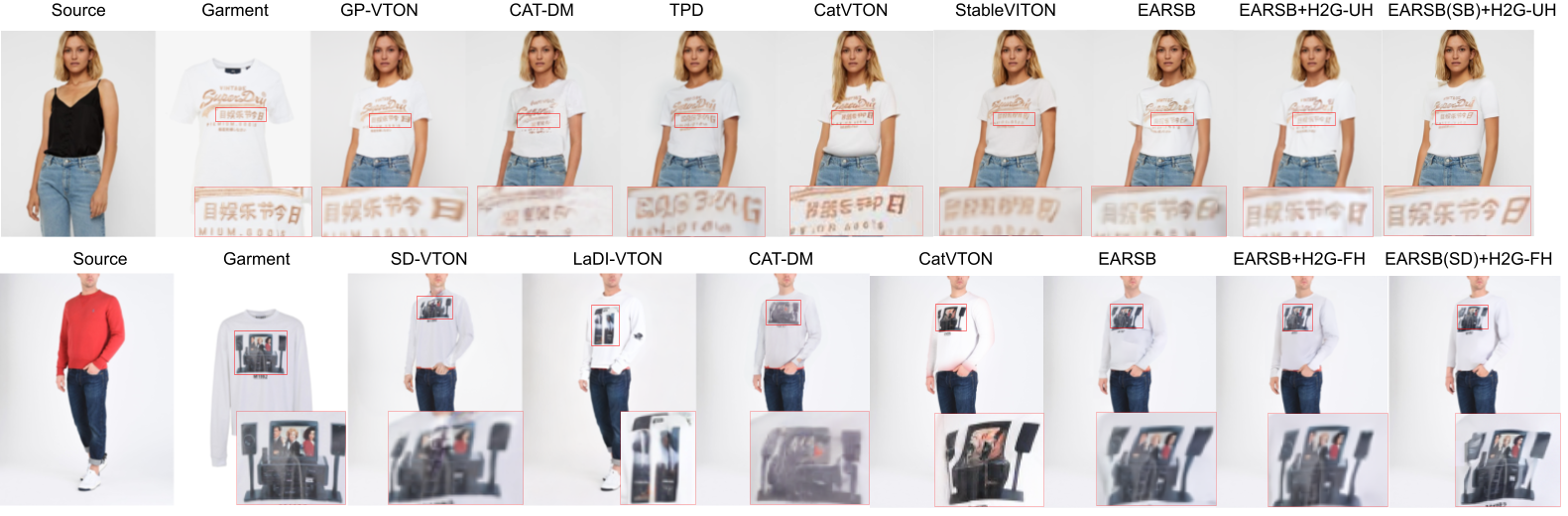}
    \vspace{-1.5em}
    \caption{Visualizations on VITON-HD (top row) and DressCode (bottom row). Our \Ours{}+H2G-UH and \Ours{}(SD)+H2G-UH better recover the intricate textures in the garment.}
    \vspace{-1.5em}
    \label{fig:vis_results}
\end{figure*}

\paragraph{User Study}
We asked Amazon MTurk workers to evaluate the texture consistency and image fidelity of synthesized images, comparing our model against GP-VTON and StableVITON. We randomly selected 100 pairs from VITON-HD to evaluate on, assigning at least 3 workers per image. Study results in \cref{tab:user} report our method is preferred at least 10\% more than the GAN-based GP-VTON and the SD-based StableVITON (59\% overall).

\paragraph{Trade-Off Between Sampling Efficiency and Image Quality}
Diffusion-generated images often show degraded quality with fewer sampling steps. %due to the large jump between noise levels. 
%\Ours{} is based on \Sch{} bridges and inherits the ability to produce reasonable results at fewer step counts as previously demonstrated in \citep{i2sb,lipmanflow}. 
In \Ours{}, the spatially adaptive noise schedule can preserve correct clothing textures in the initial image with a low noise level and only fix the erroneous parts, potentially resulting in less image quality degradation with fewer steps. In \cref{fig:ss}, while other SD-based methods have a sharp performance drop with decreasing sampling steps, \Ours{} and \Ours{}+H2G-UH show consistent performance across different sampling steps, demonstrating a better trade-off between image quality and computational efficiency.
%both \cite{i2sb} as well as closely related flow-matching works \cite{lipmanflow}

\paragraph{Qualitative Results}
\cref{fig:vis_results} gives examples of generated images from different approaches. The top row is from VITON-HD dataset and the bottom row is from DressCode-Upper. The third images in the two rows are GAN-generated results. We see that our \Ours{}+H2G-UH/FH in the last column improves the low-quality textures from the GAN-generated images, which are the distorted graphics in the center. Additional visualizations are in Supp.\ G.

\subsection{Ablations}
\label{sec:abl}
\paragraph{Synthetic Pairs Augmentation}
\cref{tab:abl_arch} incorporates H2G-UH into the training of StableVITON \citep{stableviton} and CAT-DM \citep{cat-dm} on the VITON-HD dataset to validate the effectiveness of synthetic pairs on enhancing existing diffusion methods.  We observe that that training with our synthetic H2G-UH is effective in improving most metrics.
 
 %(\ie{}, CAT-DM, StableVITON and \Ours{})
 
 \begin{table}[!t]
    \centering
    %\small
    \setlength{\tabcolsep}{1.3pt}
    \begin{tabular}{@{}lccccc@{}}
        \toprule     
         
        &\multirow{2}{*}{\shortstack[l]{Data \\ Aug.}}
        &\multicolumn{1}{c}{Unpaired} 
        & \multicolumn{3}{c}{Paired}\cr 
        \cmidrule(r){3-3} \cmidrule(r){4-6} 
        && FID$\downarrow$  & FID$\downarrow$ & SSIM$\uparrow$ & LPIPS$\downarrow$ \cr
        \midrule
        CAT-DM \citep{cat-dm} & None &8.56 &5.90 &0.911 &0.067\cr
        %&\quad +$W(\text{H2G-UH})$ &Warped &8.73  &6.12 &0.906 &0.071\cr
        CAT-DM & H2G-UH & \underline{8.36} & \underline{5.67}  & \underline{0.913} & \underline{0.063}\cr
        
        StableVITON~\citep{stableviton} & None  &8.25 &5.15  &0.917 &0.056\cr
        %&\quad +$W(\text{H2G-UH})$ & Warped &8.47 &5.32  &0.910 &0.060\cr
        StableVITON & H2G-UH & \underline{8.17} & \underline{5.04}  & \underline{0.919} & \underline{0.054}\cr

        \Ours{}(SD) &H2G-UH &\textbf{8.04} &\textbf{4.90} &\textbf{0.925} &\textbf{0.053} \cr
 \bottomrule
    \end{tabular}
    \vspace{-.5em}
    \caption{Comparing the effect of our H2G-UH data augmentation approach on VITON-HD. We bold the best overall results and underline the best results for a single model with and without H2G-UH. Each method uses the number of sampling steps from their paper: 2 for CAT-DM, 50 for StableVITON, and 25 for our own EARSB.  We find H2G-UH consistently boosts performance.}
    \label{tab:abl_arch}
    %\vspace{-1em}
\end{table}

 \begin{table}[!t]
    \centering
    %\small
    \setlength{\tabcolsep}{2pt}
    \begin{tabular}{@{}lcccc@{}}
        \toprule     
         
        &\multicolumn{1}{c}{Unpaired} 
        & \multicolumn{3}{c}{Paired}\cr 
        \cmidrule(r){2-2} \cmidrule(r){3-5} 
        & FID$\downarrow$  & FID$\downarrow$ & SSIM$\uparrow$ & LPIPS$\downarrow$ \cr    
        \midrule
        
        None  &8.42  &5.25  &0.918 &0.059 \cr
         $W(\text{H2G-UH})$ &8.68  &5.44 &0.909 &0.063\cr
        plain H2G-UH &9.64 &6.52  &0.902 &0.073 \cr%
        pre.\ H2G-UH &8.35  &5.18 &0.918 &0.059 \cr
         H2G-UH &\textbf{8.26} & \textbf{5.14} & \textbf{0.919} & \textbf{0.058} \cr
         
        % \midrule

        % \textbf{(c)}
        % &Inpaint &9.26  &6.33 &0.909 &0.068 \cr
        % &\Ours{} (w.o. $M$) &9.21  &6.27 &0.912 &0.061\cr
        % &\Ours{} (rand($M$)) &9.13  &6.55 &0.902 &0.071\cr        
        % &\Ours{} (w.o. CG)  &8.48 &5.32 &0.918 &0.059 \cr        
        
        \bottomrule
    \end{tabular}
    \vspace{-.5em}
    \caption{Ablations of our H2G-UH augmentation on VITON-HD. Specifically, \textbf{pre.\ H2G-UH} is pretrained using the synthetic pairs and finetuned on real data, \textbf{$W(\text{H2G-UH})$} replaces the synthetic garment in each pair with a randomly warped version of the clothing cropped from the real human image, \textbf{plain H2G-UH} is trained using the mixed distribution of the real and synthetic pairs \emph{without} the augmentation label identifying them, and \textbf{H2G-UH} uses the mixed data \emph{with} the identifying label.  See \cref{sec:abl} for discussion.}
    \label{tab:abl_aug}
    %\vspace{-0.5em}
\end{table}

 \begin{table}[!t]
    \centering
    %\small
    \setlength{\tabcolsep}{2.pt}
    \begin{tabular}{@{}lcccc@{}}
        \toprule     
         
        & \multicolumn{1}{c}{Unpaired} 
        & \multicolumn{3}{c}{Paired}\cr 
        \cmidrule(r){2-2} \cmidrule(r){3-5} 
        & FID$\downarrow$  & FID$\downarrow$ & SSIM$\uparrow$ & LPIPS$\downarrow$ \cr
        \midrule
        Inpainting &9.26  &6.33 &0.909 &0.068 \cr
        \Ours{} (w.o.\ $M$) &9.21  &6.27 &0.912 &0.061\cr
        \Ours{} (rand($M$)) &9.13  &6.55 &0.902 &0.071\cr        
        \Ours{} (w.o.\ CG)  &8.48 &5.32 &\textbf{0.918} &\textbf{0.059} \cr
        \Ours{} (full) &\textbf{8.42}  &\textbf{5.25}  &\textbf{0.918} &\textbf{0.059} \cr
        
        \bottomrule
    \end{tabular}
    \vspace{-.5em}
    \caption{Comparing noise scheduling strategies on VITON-HD. We include a simple inpainting baseline, not using the error map during training (w.o.\ $M$), a random error map (rand($M$)), and removing classifier guidance (w.o.\ CG).  These results demonstrate the importance of using a meaningful error map.}
    \label{tab:abl2}
    %\vspace{-1em}
\end{table}

%To determine the importance of the canonical product-view projection, we replace the synthetic garment in each pair with a randomly warped version of the clothing cropped from the real human image, denoted as +$W(\text{H2G-UH})$. The results show that while +$W(\text{H2G-UH})$ hinders the performance of all baseline approaches, incorporating H2G-UH improves most metrics. This demonstrates that the synthetic product-view contributes significantly to the observed improvements.
\cref{tab:abl_aug} explores different ways of incorporating the synthetic data H2G-UH during the training of \Ours{}. 
%Further, we see the effect of conditioning on the augmentation label, as removing it in \Ours{}+plain H2G-UH greatly degrades the image quality and causes a significant drop in all metrics.
We find that using a randomly warped version of the clothing ($W(\text{H2G-UH})$) hinders performance, demonstrating the importance of the synthetic product-view.
Additionally, the poor \emph{plain H2G-UH} results indicate %the indicator for generated samples shows that 
the presence of a synthetic-real domain gap when using H2G-UH. Inspired by \citep{QraitemFFR2024}, one way to address this issue is by pretraining  on the synthetic data and finetuning on real samples (see pre.\ H2G-UH). However, we find it most effective to condition on a synthetic data indicator while training on mixed data.

%We find using H2G-UH during training provides a small but consistent boost over only using it during pretraining (pre.). 

%However, a flag identifying synthetic samples helps our approach benefit from share parameters while minimizing the potential for synthetic data distribution to leaking directly into the modeling of the real data distribution (last line of \cref{tab:abl_aug}).

\paragraph{Error-Aware Noise Schedule}
\cref{tab:abl2} explores the importance of the error-aware noise schedule described in \cref{sec:pipeline}, where the error map adapts the noise distribution according to the quality of the image patches in the initial image $x_1$. This adaptive approach contrasts with a uniform Gaussian noise application across all locations, which would reduce our model to I$^2$SB. We consider three baselines: \textbf{Inpainting} regenerates rather than refines erroneous regions (where the mean confidence $M$ of an image patch containing an error is greater than 0.5), \textbf{w.o.\ $M$} removes the error map during training, and \textbf{rand($M$)} indicates a random error map.  As shown in \cref{tab:abl2}, \Ours{} outperforms all these baselines, underscoring the importance of a meaningful error map in precisely locating and enhancing targeted regions. Additionally, the slight performance degradation observed when removing the classifier guidance (\Ours{}(w.o.\ CG)) suggests that the error map employed in our classifier guidance also contributes to overall image quality improvement. Collectively, these findings highlight the crucial role of our adaptive noise schedule in achieving superior results.

%As shown in \cref{tab:abl2}, removing the error map during training (\Ours{}(w.o.\ $M$)) results in a substantial decline across all metrics. 
%In another setting, instead of learning the refinement with adaptive noise distribution, we train an inpainting model Inpaint to directly regenerate the erroneous regions in $x_0$ using the same model architecture. In this experiment, the image patch in $x_0$ is defined to be erroneous if its mean confidence in is greater than 0.5. In \cref{tab:abl2}, the inpainting model Inpaint shows degraded performance, demonstrating the importance of learning the error-to-ground-truth refinement in our model.
%Furthermore, we explored using a random error map during the sampling process (\Ours{}()), which also leads to diminished performance. These results  

 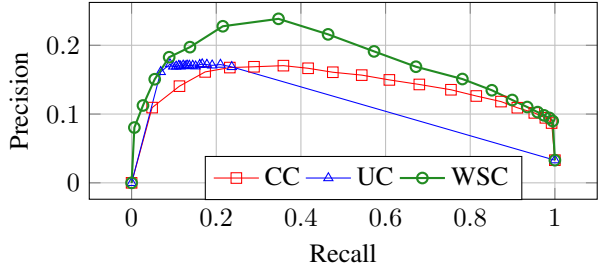
\begin{figure}
    \definecolor{forestgreen}{RGB}{34, 139, 34}
\begin{tikzpicture}
    \begin{axis}[
        xlabel={Recall},
        ylabel={Precision},
        legend style={at={(0.54,0.22)}, anchor=north, legend columns=-1},
        grid=major,
        xlabel near ticks,
        ylabel near ticks,
        width=\columnwidth, height=4.2cm
    ]
        % Plot for the first CSV file
        \addplot[
            color=red,
            mark=square
        ] 
        table [col sep=comma, x index=0, y index=1] {figures/data/hc_recall_precision.csv};
        \addlegendentry{CC}

        % Plot for the second CSV file
        \addplot[
            color=blue,
            mark=triangle
        ] 
        table [col sep=comma, x index=0, y index=1] {figures/data/uc_recall_precision.csv};
        \addlegendentry{UC}

        % Plot for the third CSV file
        \addplot[
            color=forestgreen,
            thick,
            mark=o
        ] 
        table [col sep=comma, x index=0, y index=1] {figures/data/wsc_recall_precision.csv};
        \addlegendentry{WSC}
    \end{axis}
\end{tikzpicture}
    \vspace{-0.5em}
    \caption{The precision-recall curve for retrieving annotated errors at the pixel level, comparing our WSC to two fully unsupervised baselines (UC, CC). WSC performs best at retrieving generation artifacts at a nominal labeling cost.}
    %\vspace{-1em}
\label{fig:prc}
\end{figure}

\paragraph{Weakly-Supervised error Classifier (WSC)}
Our weakly supervised classifier from \cref{sec:classifier} highlights low-quality regions in the initial image $x_1$ with only a few hours of labeling. To validate its effectiveness, we train two ablations of our WSC: the Unsupervised Classifier (UC) that only uses image labels (\ie{}, fake or real), and the Fake/Real Composite Classifier (CC). CC uses both image fake/real labels as well as fake region-level labels which are created by compositing real image patches and fake image patches. The compositing is a fully automatic alternative to manual labeling that provides patch-level labels. We annotated 100 images in the test set to validate their effectiveness. \cref{fig:prc} shows the pixel-level precision-recall curve for retrieving annotated artifact pixels within the bounding boxes using the classifiers' confidence maps. It is clear that weak supervision remains an incredibly cost-effective approach.

% \paragraph{Effect of Quality of the Initial Image}
% %We chose the GAN-based try-on model to obtain the initial image $x_1$ due to the GAN's efficiency in generating samples. Better GAN models will give us $x_1$ of higher image quality. F
% Following CAT-DM \citep{cat-dm}, we use different try-on GAN models, including HR-VTON \citep{hrviton}, SD-VTON \citep{sd-viton} and GP-VTON \citep{gp-vton} at sampling to see how the quality of $x_1$ affects the sampled images at 25 sampling steps. See results in Appendix D, Tab. 4. We observe that: a) our \Ours{} can refine the GAN-generated image over the GAN baseline; b) the quality of the initial image $x_1$ is positively correlated with the quality of the sampled $\hat{x}_0$; c) our model achieves higher gains over CAT-DM, which also tries to refine the GAN-generated image but without error-aware noise schedule.

\section{Conclusion}
This paper addresses two shortcomings of prior work on virtual try-on. First, we address the limited data availability by introducing a human-to-garment model that generates (human, synthetic garment) pairs from a single image of a clothed individual. Second, we propose a refinement model \Ours{} that surgically targets localized generation errors from the output of a prior model. \Ours{} improves the low-quality region of an initially generated image based on a spatially-varying noise schedule that targets known artifacts. Experiments on two benchmark datasets demonstrate that our synthetic data augmentation improves the performance of prior work and that \Ours{} enhances the overall image quality.

%In \Ours{}, a weakly-supervised error classifier is introduced to identify localized regions to refine. These identified regions are incorporated into the noise schedule of SB with the error classifier's error map. 

\noindent\textbf{Acknowledgments} This material is based upon work supported, in part, by DARPA under agreement number HR00112020054 and the National Science Foundation under Grant No.\ DBI-2134696. Any opinions, findings, and conclusions or recommendations expressed in this material are those of the author(s) and do not necessarily reflect the views of the supporting agencies.


{
    \small
    \bibliographystyle{ieeenat_fullname}
    \bibliography{main}

\begin{thebibliography}{41}
\providecommand{\natexlab}[1]{#1}
\providecommand{\url}[1]{\texttt{#1}}
\expandafter\ifx\csname urlstyle\endcsname\relax
  \providecommand{\doi}[1]{doi: #1}\else
  \providecommand{\doi}{doi: \begingroup \urlstyle{rm}\Url}\fi

\bibitem[Alemohammad et~al.(2024)Alemohammad, Humayun, Agarwal, Collomosse, and Baraniuk]{alemohammad2024self}
Sina Alemohammad, Ahmed~Imtiaz Humayun, Shruti Agarwal, John Collomosse, and Richard Baraniuk.
\newblock Self-improving diffusion models with synthetic data.
\newblock \emph{arXiv preprint arXiv:2408.16333}, 2024.

\bibitem[Balaji et~al.(2022)Balaji, Nah, Huang, Vahdat, Song, Zhang, Kreis, Aittala, Aila, Laine, Catanzaro, Karras, and Liu]{balaji2022eDiff-I}
Yogesh Balaji, Seungjun Nah, Xun Huang, Arash Vahdat, Jiaming Song, Qinsheng Zhang, Karsten Kreis, Miika Aittala, Timo Aila, Samuli Laine, Bryan Catanzaro, Tero Karras, and Ming-Yu Liu.
\newblock {eDiff-I}: Text-to-image diffusion models with ensemble of expert denoisers.
\newblock \emph{arXiv preprint arXiv:2211.01324}, 2022.

\bibitem[Baldrati et~al.(2023)Baldrati, Morelli, Cartella, Cornia, Bertini, and Cucchiara]{baldrati2023multimodal}
Alberto Baldrati, Davide Morelli, Giuseppe Cartella, Marcella Cornia, Marco Bertini, and Rita Cucchiara.
\newblock Multimodal garment designer: Human-centric latent diffusion models for fashion image editing.
\newblock In \emph{CVPR}, 2023.

\bibitem[Bi{\'n}kowski et~al.(2018)Bi{\'n}kowski, Sutherland, Arbel, and Gretton]{kid}
Miko{\l}aj Bi{\'n}kowski, Danica~J Sutherland, Michael Arbel, and Arthur Gretton.
\newblock Demystifying {MMD GAN}s.
\newblock In \emph{ICLR}, 2018.

\bibitem[Brooks et~al.(2023)Brooks, Holynski, and Efros]{brooks2023instructpix2pix}
Tim Brooks, Aleksander Holynski, and Alexei~A Efros.
\newblock Instructpix2pix: Learning to follow image editing instructions.
\newblock In \emph{CVPR}, 2023.

\bibitem[Choi et~al.(2024)Choi, Kwak, Lee, Choi, and Shin]{idmvton}
Yisol Choi, Sangkyung Kwak, Kyungmin Lee, Hyungwon Choi, and Jinwoo Shin.
\newblock Improving diffusion models for virtual try-on.
\newblock In \emph{ECCV}, 2024.

\bibitem[Chong et~al.(2024)Chong, Dong, Li, Zhang, Zhang, Zhang, Zhao, and Liang]{catvton}
Zheng Chong, Xiao Dong, Haoxiang Li, Shiyue Zhang, Wenqing Zhang, Xujie Zhang, Hanqing Zhao, and Xiaodan Liang.
\newblock Catvton: Concatenation is all you need for virtual try-on with diffusion models.
\newblock \emph{arXiv preprint arXiv:2407.15886}, 2024.

\bibitem[Chung et~al.(2023)Chung, Kim, Mccann, Klasky, and Ye]{chung2023diffusion}
Hyungjin Chung, Jeongsol Kim, Michael~T Mccann, Marc~L Klasky, and Jong~Chul Ye.
\newblock Diffusion posterior sampling for general noisy inverse problems.
\newblock In \emph{ICLR}, 2023.

\bibitem[Cui et~al.(2023)Cui, He, Xiang, and Toisoul]{cui2023learning}
Aiyu Cui, Sen He, Tao Xiang, and Antoine Toisoul.
\newblock Learning garment densepose for robust warping in virtual try-on.
\newblock \emph{arXiv preprint arXiv:2303.17688}, 2023.

\bibitem[Dablain et~al.(2022)Dablain, Krawczyk, and Chawla]{dablain2022deepsmote}
Damien Dablain, Bartosz Krawczyk, and Nitesh~V Chawla.
\newblock Deepsmote: Fusing deep learning and smote for imbalanced data.
\newblock \emph{IEEE Transactions on Neural Networks and Learning Systems}, 34\penalty0 (9):\penalty0 6390--6404, 2022.

\bibitem[Dhariwal and Nichol(2021)]{diff_beat_gan}
Prafulla Dhariwal and Alexander Nichol.
\newblock Diffusion models beat gans on image synthesis.
\newblock \emph{NeurIPS}, 2021.

\bibitem[Fu et~al.(2022)Fu, Li, Jiang, Lin, Qian, Loy, Wu, and Liu]{sshq}
Jianglin Fu, Shikai Li, Yuming Jiang, Kwan-Yee Lin, Chen Qian, Chen~Change Loy, Wayne Wu, and Ziwei Liu.
\newblock Stylegan-human: A data-centric odyssey of human generation.
\newblock In \emph{European Conference on Computer Vision}, 2022.

\bibitem[Gao et~al.(2023)Gao, Liu, Zeng, Xu, Li, Luo, Liu, Zhen, and Zhang]{ddim}
Sicheng Gao, Xuhui Liu, Bohan Zeng, Sheng Xu, Yanjing Li, Xiaoyan Luo, Jianzhuang Liu, Xiantong Zhen, and Baochang Zhang.
\newblock Implicit diffusion models for continuous super-resolution.
\newblock In \emph{CVPR}, 2023.

\bibitem[G{\"u}ler et~al.(2018)G{\"u}ler, Neverova, and Kokkinos]{guler2018densepose}
R{\i}za~Alp G{\"u}ler, Natalia Neverova, and Iasonas Kokkinos.
\newblock Densepose: Dense human pose estimation in the wild.
\newblock In \emph{CVPR}, 2018.

\bibitem[Han et~al.(2019)Han, Hu, Huang, and Scott]{clothflow}
Xintong Han, Xiaojun Hu, Weilin Huang, and Matthew~R Scott.
\newblock Clothflow: A flow-based model for clothed person generation.
\newblock In \emph{CVPR}, 2019.

\bibitem[Heusel et~al.(2017)Heusel, Ramsauer, Unterthiner, Nessler, and Hochreiter]{fid}
Martin Heusel, Hubert Ramsauer, Thomas Unterthiner, Bernhard Nessler, and Sepp Hochreiter.
\newblock {GAN}s trained by a two time-scale update rule converge to a local nash equilibrium.
\newblock In \emph{NeurIPS}, 2017.

\bibitem[Jun et~al.(2020)Jun, Child, Chen, Schulman, Ramesh, Radford, and Sutskever]{jun2020distribution}
Heewoo Jun, Rewon Child, Mark Chen, John Schulman, Aditya Ramesh, Alec Radford, and Ilya Sutskever.
\newblock Distribution augmentation for generative modeling.
\newblock In \emph{ICML}, 2020.

\bibitem[Karras et~al.(2020)Karras, Aittala, Hellsten, Laine, Lehtinen, and Aila]{karras2020training}
Tero Karras, Miika Aittala, Janne Hellsten, Samuli Laine, Jaakko Lehtinen, and Timo Aila.
\newblock Training generative adversarial networks with limited data.
\newblock \emph{NeurIPS}, 2020.

\bibitem[Kim et~al.(2024)Kim, Gu, Park, Park, and Choo]{stableviton}
Jeongho Kim, Guojung Gu, Minho Park, Sunghyun Park, and Jaegul Choo.
\newblock {StableVITON}: Learning semantic correspondence with latent diffusion model for virtual try-on.
\newblock In \emph{CVPR}, 2024.

\bibitem[Kumar et~al.(2022)Kumar, Raghunathan, Jones, Ma, and Liang]{finetune}
Ananya Kumar, Aditi Raghunathan, Robbie~Matthew Jones, Tengyu Ma, and Percy Liang.
\newblock Fine-tuning can distort pretrained features and underperform out-of-distribution.
\newblock In \emph{ICLR}, 2022.

\bibitem[Lee et~al.(2022)Lee, Gu, Park, Choi, and Choo]{hrviton}
Sangyun Lee, Gyojung Gu, Sunghyun Park, Seunghwan Choi, and Jaegul Choo.
\newblock High-resolution virtual try-on with misalignment and occlusion-handled conditions.
\newblock In \emph{ECCV}, 2022.

\bibitem[Li et~al.(2024)Li, Liu, Singh, Wang, Zhang, Plummer, and Lin]{unihuman}
Nannan Li, Qing Liu, Krishna~Kumar Singh, Yilin Wang, Jianming Zhang, Bryan~A Plummer, and Zhe Lin.
\newblock {UniHuman}: A unified model for editing human images in the wild.
\newblock In \emph{CVPR}, 2024.

\bibitem[Li et~al.(2023)Li, Wei, Yin, Ma, and Kot]{keypoints-tryon}
Zhi Li, Pengfei Wei, Xiang Yin, Zejun Ma, and Alex~C Kot.
\newblock Virtual try-on with pose-garment keypoints guided inpainting.
\newblock In \emph{ICCV}, 2023.

\bibitem[Liu et~al.(2023)Liu, Vahdat, Huang, Theodorou, Nie, and Anandkumar]{i2sb}
Guan-Horng Liu, Arash Vahdat, De-An Huang, Evangelos~A Theodorou, Weili Nie, and Anima Anandkumar.
\newblock I2sb: image-to-image schr{\"o}dinger bridge.
\newblock In \emph{ICML}, 2023.

\bibitem[Liu et~al.(2016)Liu, Luo, Qiu, Wang, and Tang]{deepfashion}
Ziwei Liu, Ping Luo, Shi Qiu, Xiaogang Wang, and Xiaoou Tang.
\newblock Deepfashion: Powering robust clothes recognition and retrieval with rich annotations.
\newblock In \emph{CVPR}, 2016.

\bibitem[Morelli et~al.(2022)Morelli, Fincato, Cornia, Landi, Cesari, and Cucchiara]{dresscode}
Davide Morelli, Matteo Fincato, Marcella Cornia, Federico Landi, Fabio Cesari, and Rita Cucchiara.
\newblock Dress code: High-resolution multi-category virtual try-on.
\newblock In \emph{CVPR}, 2022.

\bibitem[Morelli et~al.(2023)Morelli, Baldrati, Cartella, Cornia, Bertini, and Cucchiara]{ladiviton}
Davide Morelli, Alberto Baldrati, Giuseppe Cartella, Marcella Cornia, Marco Bertini, and Rita Cucchiara.
\newblock {LaDI-VTON: Latent Diffusion Textual-Inversion Enhanced Virtual Try-On}.
\newblock In \emph{ACM Multimedia}, 2023.

\bibitem[Ning et~al.(2024)Ning, Wang, Qin, Jin, Wang, and Han]{ning2024picture}
Shuliang Ning, Duomin Wang, Yipeng Qin, Zirong Jin, Baoyuan Wang, and Xiaoguang Han.
\newblock Picture: Photorealistic virtual try-on from unconstrained designs.
\newblock In \emph{CVPR}, 2024.

\bibitem[Qraitem et~al.(2024)Qraitem, Saenko, and Plummer]{QraitemFFR2024}
Maan Qraitem, Kate Saenko, and Bryan~A. Plummer.
\newblock From fake to real: Pretraining on balanced synthetic images to prevent spurious correlations in image recognition.
\newblock In \emph{The European Conference on Computer Vision (ECCV)}, 2024.

\bibitem[Rombach et~al.(2022)Rombach, Blattmann, Lorenz, Esser, and Ommer]{SD}
Robin Rombach, Andreas Blattmann, Dominik Lorenz, Patrick Esser, and Bj{\"o}rn Ommer.
\newblock High-resolution image synthesis with latent diffusion models.
\newblock In \emph{CVPR}, 2022.

\bibitem[Shim et~al.(2024)Shim, Chung, and Heo]{sd-viton}
Sang-Heon Shim, Jiwoo Chung, and Jae-Pil Heo.
\newblock Towards squeezing-averse virtual try-on via sequential deformation.
\newblock In \emph{AAAI}, 2024.

\bibitem[Shivashankar and Miller(2023)]{semantic_aug}
C Shivashankar and Shane Miller.
\newblock Semantic data augmentation with generative models.
\newblock In \emph{CVPR}, 2023.

\bibitem[Song and Ermon(2019)]{score_match}
Yang Song and Stefano Ermon.
\newblock Generative modeling by estimating gradients of the data distribution.
\newblock \emph{NeurIPS}, 2019.

\bibitem[Wang et~al.(2004)Wang, Bovik, Sheikh, and Simoncelli]{ssim}
Zhou Wang, Alan~C Bovik, Hamid~R Sheikh, and Eero~P Simoncelli.
\newblock Image quality assessment: from error visibility to structural similarity.
\newblock \emph{IEEE Transactions on Image Processing}, 13\penalty0 (4):\penalty0 600--612, 2004.

\bibitem[Wasserman et~al.(2024)Wasserman, Rotstein, Ganz, and Kimmel]{wasserman2024paint}
Navve Wasserman, Noam Rotstein, Roy Ganz, and Ron Kimmel.
\newblock Paint by inpaint: Learning to add image objects by removing them first.
\newblock \emph{arXiv preprint arXiv:2404.18212}, 2024.

\bibitem[Xie et~al.(2021)Xie, Huang, Zhao, Dong, Kampffmeyer, and Liang]{upt}
Zhenyu Xie, Zaiyu Huang, Fuwei Zhao, Haoye Dong, Michael Kampffmeyer, and Xiaodan Liang.
\newblock Towards scalable unpaired virtual try-on via patch-routed spatially-adaptive gan.
\newblock \emph{NeurIPS}, 2021.

\bibitem[Xie et~al.(2023)Xie, Huang, Dong, Zhao, Dong, Zhang, Zhu, and Liang]{gp-vton}
Zhenyu Xie, Zaiyu Huang, Xin Dong, Fuwei Zhao, Haoye Dong, Xijin Zhang, Feida Zhu, and Xiaodan Liang.
\newblock {GP-VTON}: Towards general purpose virtual try-on via collaborative local-flow global-parsing learning.
\newblock In \emph{CVPR}, 2023.

\bibitem[Yang et~al.(2024)Yang, Ding, Hong, Huang, Tao, and Xu]{tpd}
Xu Yang, Changxing Ding, Zhibin Hong, Junhao Huang, Jin Tao, and Xiangmin Xu.
\newblock Texture-preserving diffusion models for high-fidelity virtual try-on.
\newblock In \emph{CVPR}, 2024.

\bibitem[Zeng et~al.(2024)Zeng, Song, Nie, Tian, Wang, and Liu]{cat-dm}
Jianhao Zeng, Dan Song, Weizhi Nie, Hongshuo Tian, Tongtong Wang, and An-An Liu.
\newblock Cat-dm: Controllable accelerated virtual try-on with diffusion model.
\newblock In \emph{CVPR}, 2024.

\bibitem[Zhang et~al.(2018)Zhang, Isola, Efros, Shechtman, and Wang]{lpips}
Richard Zhang, Phillip Isola, Alexei~A Efros, Eli Shechtman, and Oliver Wang.
\newblock The unreasonable effectiveness of deep features as a perceptual metric.
\newblock In \emph{CVPR}, 2018.

\bibitem[Zhu et~al.(2023)Zhu, Yang, Zhu, Reda, Chan, Saharia, Norouzi, and Kemelmacher-Shlizerman]{tryondiffusion}
Luyang Zhu, Dawei Yang, Tyler Zhu, Fitsum Reda, William Chan, Chitwan Saharia, Mohammad Norouzi, and Ira Kemelmacher-Shlizerman.
\newblock {TryOnDiffusion}: A tale of two unets.
\newblock In \emph{CVPR}, 2023.

\end{thebibliography}
}

\clearpage

\setcounter{figure}{7}
\setcounter{table}{5}

\appendix
\appendixpage

\begin{table}[h]
    \centering
    \begin{tabular}{cc} 
    Parameter &Value \cr
    \toprule
    Batch Ratio of Synthetic Data &15\% \cr
    Batch Size &32 \cr
    Image Size & 512x512 \cr
    \#Model Parameters    &102.6M \cr
    Learning Rate &$10^{-4}$ \cr
    \#Training Iterations &200K \cr
    \#Finetuning Iterations &100K \cr
    \bottomrule
    \end{tabular}
    \caption{Implementation details of \Ours{}+H2G-UH/FH.}
    \label{tab:implement}
\end{table}

\section{Implementations Details}
For generating the initial image $x_1$ in our \Ours{} training, we employ three try-on GAN models: HR-VTON \citep{hrviton} and SD-VTON \citep{sd-viton} and GP-VTON \citep{gp-vton}. All human images are processed to maintain their aspect ratio, with the longer side resized to 512 pixels and the shorter side padded with white pixels to reach 512. During training, images undergo random shifting and flipping with a 0.2 probability. The weakly-supervised classifier is trained for 100K iterations with a batch size of 8, while the human-to-garment GAN is trained for 90K iterations with a batch size of 16. As shown in Tab. \ref{tab:implement}, \Ours{}+H2G-UH/FH is trained for 300K iterations with a batch size of 32, incorporating 15\% synthetic pairs in each batch. The first 200K iterations are trained on $t \in [0,1]$ while the following 100k iterations are finetuned on $t \in [0,0.5)$ and $t \in [0.5,1]$ respectively following \citep{balaji2022eDiff-I}. All models utilize the AdamW optimizer with a learning rate of $10^{-4}$.

%and then finetune the model on $t \in [0,0.5)$ and $t \in [0.5,1]$, respectively following \citep{balaji2022eDiff-I}

For inference, we select the GAN model that demonstrates better performance on each dataset to generate the initial image. Specifically, we employ GP-VTON \citep{gp-vton} for VITON-HD and SD-VTON \citep{sd-viton} for DressCode-Upper. During the sampling process, the guidance score in Eq. (10) is scaled by a factor of 6 and clamped to the range $[-0.3, 0.3]$.

\begin{figure}[!h]
    \centering
    \centering
        \includegraphics[width=0.9\linewidth]{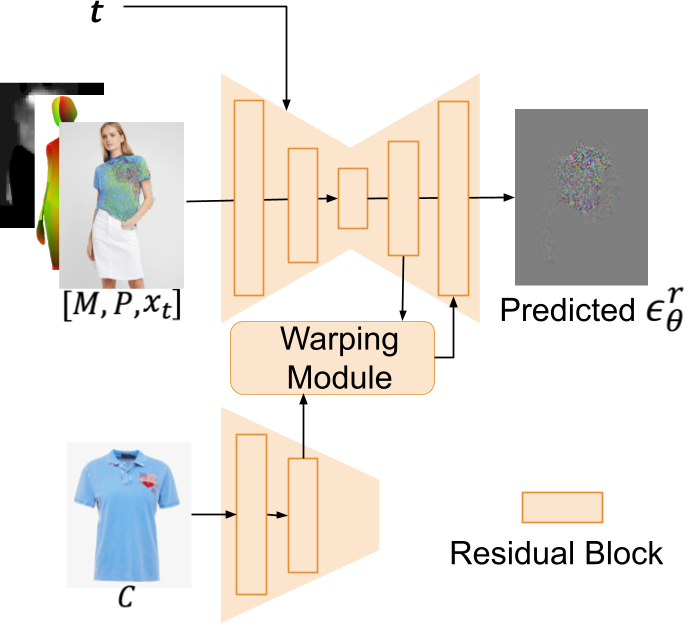}
    \caption{Architecture of our UNet in \Ours{}.}
    \label{fig:earsb_unet}
\end{figure}

\begin{figure}[!h]
    \centering
        \includegraphics[width=0.9\linewidth]{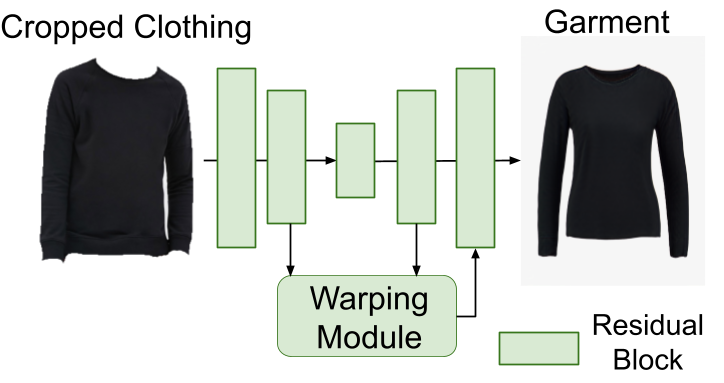}
    \caption{Architecture of our UNet in the human-to-garment model.}
    \label{fig:h2g_unet} 
    \vspace{-0.5cm}
\end{figure}
\section{UNet Architecture}
\label{sec:unet}

\begin{figure*}[!t]
     \centering
    \includegraphics[width=\textwidth]{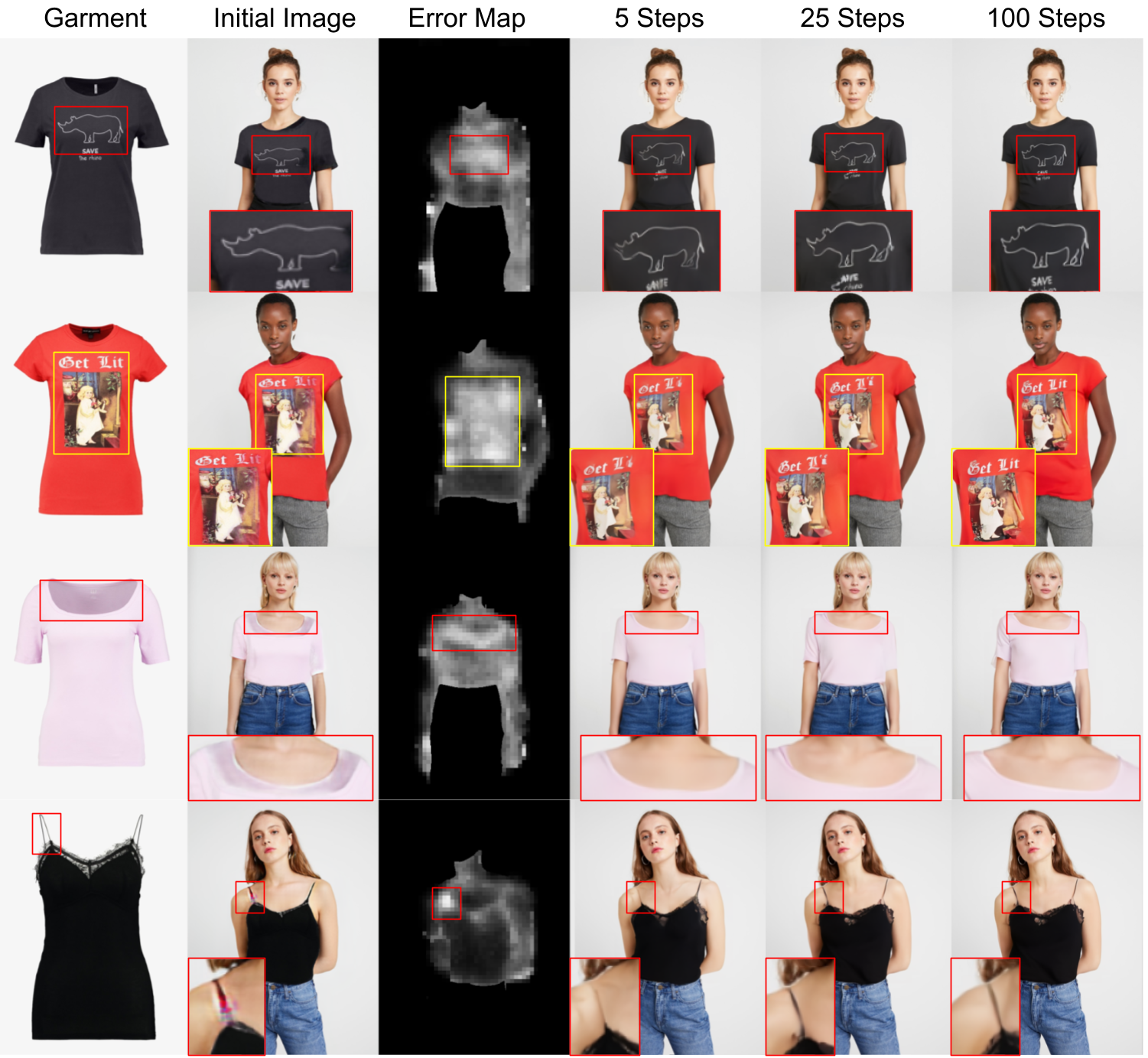}
    \caption{Results on different time steps. Our error map focuses on low-quality regions and maintains the quality of the sufficiently good regions.}
    \label{fig:vis_error}
\end{figure*}

\paragraph{\Ours{} UNet}
The UNet architecture in \Ours{} consists of residual blocks and garment warping modules. It processes the concatenation of the error map $M$, pose representation $P$, and noisy image $x_t$ to predict the noise distribution $\epsilon^r_{\theta}$ at time $t$. The UNet encoder has 21 residual blocks, with the number of channels doubling every three blocks to a maximum of 256. Similarly, the garment encoder has 21 residual blocks but reaches a maximum of 128 channels. The decoder mirrors the encoder's structure, with extra garment warping modules. As shown in Fig. \ref{fig:earsb_unet}, each of the first 15 residual blocks in the UNet decoder is followed by a convolutional warping module. These modules concatenate encoded garment features and UNet-decoded features to predict a flow-like map for spatially warping the encoded garment features. The warped features are then injected into the subsequent decoder layer via input concatenation. Following \citep{SD}, all residual blocks and flow-learning modules incorporate timestep embeddings to renormalize latent features.

\paragraph{Human-to-Garment UNet}
Our human-to-garment UNet architecture is adapted from the model proposed in \citep{clothflow}. As illustrated in Fig. \ref{fig:h2g_unet}, it shares similarities with the UNet in \Ours{}, but with two key distinctions:
a) It is not timestep-dependent and takes cropped clothing as input to generate its product-view image. 
b) The garment warping module utilizes the i$_{th}$ clothing features from both the encoder and decoder to learn a flow-like map, rather than using encoded features from the human.

\begin{figure*}[!t]
     \centering
    \includegraphics[width=0.85\textwidth]{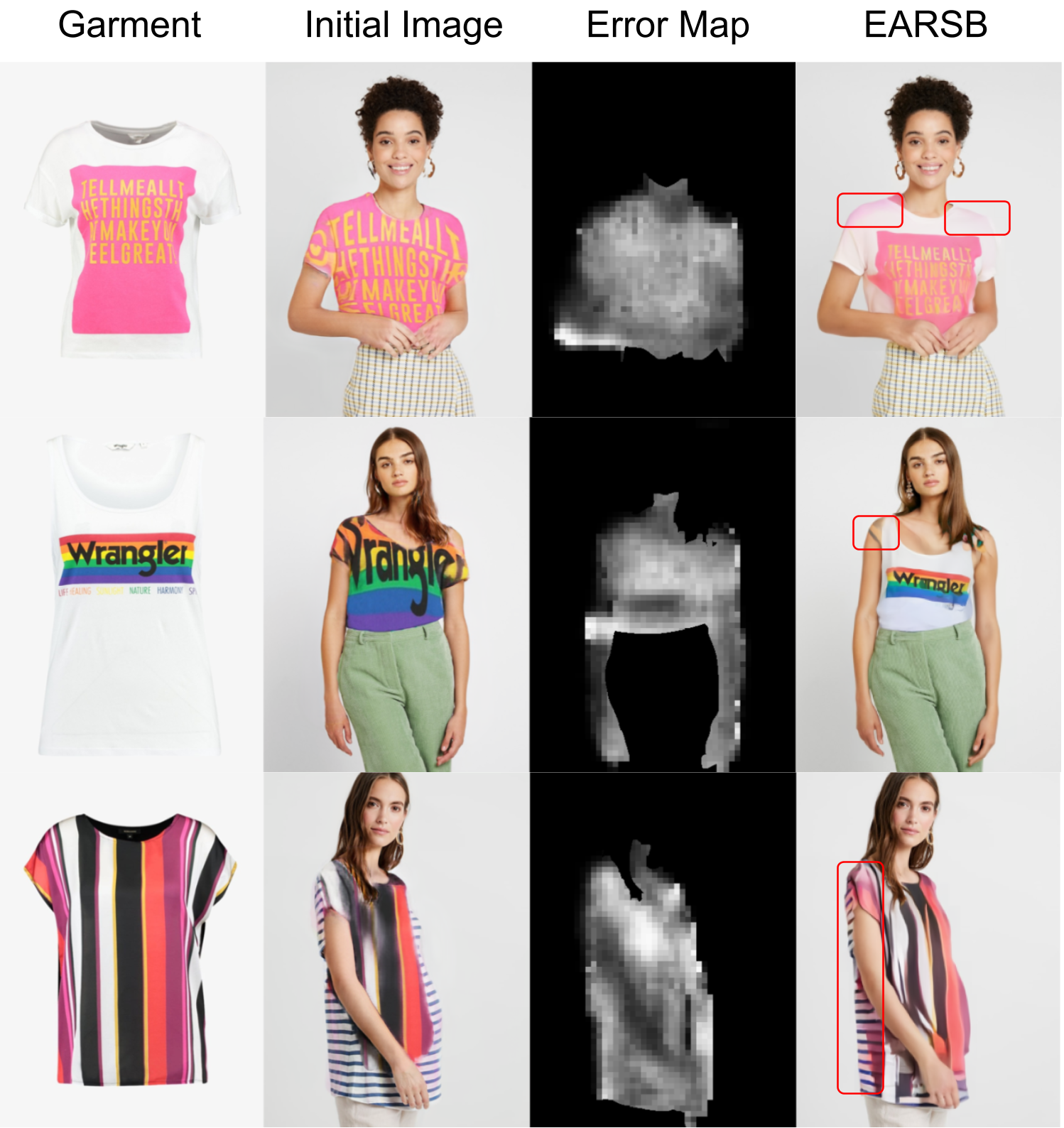}
    \caption{Failure cases on VITON-HD where the initial image has a poor-quality.}
    \label{fig:fails}
\end{figure*}

\section{Visualizing Error Maps}

Our \Ours{} focuses on fixing specific errors and therefore can save the sampling cost when initial predictions are sufficiently good. For example, in the first row of Fig. \ref{fig:vis_error}, the error map highlights the graphics and text in the initial image. This low-quality part is being refined progressively as the number of sampling steps increases from 5 to 100. At the same time, other parts that our weakly-supervised classifier believes to be sufficiently good, which are mostly the solid-color areas, are kept well regardless of the number of sampling steps. Therefore, for an initial image whose error map has almost zero values, we can choose to use fewer steps in sampling. On the contrary, for an initial image whose error map has high confidence, we should assign more sampling steps to it to improve the image quality.

\begin{table}[!t]
    \centering
    \small
    \setlength{\tabcolsep}{2pt}
    \begin{tabular}{@{}lccc@{}}
        \toprule     
      &HR-VTON \citep{hrviton} &SD-VTON \citep{sd-viton} &GP-VTON \citep{gp-vton} \cr
      \midrule
      Baseline &10.75 &9.05 &8.61\cr
      CAT-DM \citep{cat-dm}  &10.03 &8.76  &8.55\cr
      \Ours{} &\textbf{9.11} &\textbf{8.69} &\textbf{8.42}\cr
        
        \bottomrule
    \end{tabular}
    \caption{FID scores of using different try-on GAN models to generate the initial image under the unpaired setting. }
    \label{tab:three_gans}
\end{table}
\section{Ablations on the Quality of the Initial Image}
In Tab. \ref{tab:three_gans} we include the FID results of using different try-on GAN models to generate the initial image under the unpaired setting. Baseline means the GAN baseline. We can draw three conclusions from the results: a) our \Ours{} can refine the GAN-generated image over the GAN baseline; b) the quality of the initial image $x_1$ is positively correlated with the quality of the sampled $\hat{x}_0$; c) our model achieves higher gains over CAT-DM, which also tries to refine the GAN-generated image but without error-aware noise schedule.

\begin{figure}[!t]
    \centering
    \includegraphics[width=\linewidth]{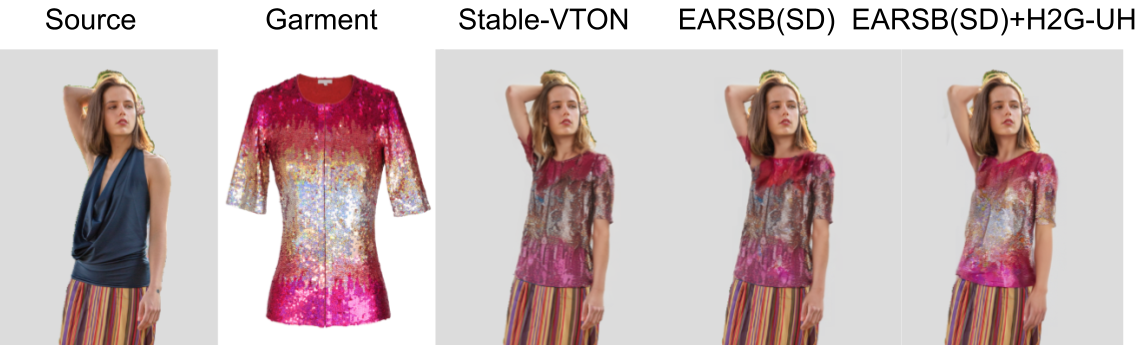}
    \caption{Visualization of the generated images in WVTON. }
    \label{fig:wvton}
\end{figure}
\begin{table}[!t]
    \centering
    \footnotesize
    \begin{tabular}{ccc}
       &FID$\downarrow$ &KID$\downarrow$ \cr
       \toprule
       Stable-VTON \citep{stableviton} &131.76 &2.10 \cr
       EARSB(SD) &127.15 &1.67\cr
       EARSB(SD) +H2G-UH &\textbf{120.29} &\textbf{1.18} \cr
       \bottomrule
    \end{tabular}
    \caption{Results on out-of-domain test set WVTON under the unpaired setting. All image background is removed for evaluation.}
    \label{tab:ood}
\end{table}

\section{Results on In-the-Wild Dataset}
We ran our data-augmented EARSB on the Out-of-Domain test set WVTON \citep{unihuman} under the unpaired setting and removed the image background for evaluation. In \cref{tab:ood}, we observe a 7 point gain in FID, showing its good generalization ability. \cref{fig:wvton} also shows that EARSB(SD)+H2G-UH better recovers the clothing patterns. 

\section{Limitations}
While our human-to-garment model can effectively generate synthetic paired data for try-on training augmentation, it has some imperfections. The overall quality of synthetic garments is regulated by our filtering criteria (Sec. 3.2), yet minor texture deformations occasionally occur. For instance, in Fig. \ref{fig:more_vis_h2g_uh}, the second pair of the first row shows a misaligned shirt placket in the synthetic garment. This limitation stems partly from the fact that our model is trained in the image domain which lacks 3D information. A potential solution is to utilize DensePose representations extracted from the garment as in \citep{cui2023learning}.
%Additionally, the model lacks prior knowledge to infer the correct relative positions of small clothing elements, such as centered plackets.

A key constraint of our \Ours{} is its refinement-based nature, which makes the generated image dependent on the initial image. We assume that the initial image from a try-on GAN model is of reasonable quality, requiring only partial refinement. Consequently, if the initial image is of very poor quality, our refinement process cannot completely erase and regenerate an entirely new, unrelated image. Fig. \ref{fig:fails} illustrates this limitation: in the first row, the initial image severely mismatches the white shirt with pink graphics. With \Ours{} refinement, while the shirt is correctly re-warped, color residuals from the initial image persist around the shoulder area.

\begin{table}[!t]
    \centering
    \begin{tabular}{c|cccc}
    &FID &KID &SSIM &LPIPS \cr
    \toprule
    VITON-HD  &14.81 &0.42 &0.849 &0.229 \cr
    DressCode-Upper &18.92 &0.59 &0.832 &0.257   \cr
    \bottomrule
    \end{tabular}
    \caption{Human-to-Garment results under 1024x1024 image resolution.}
    \label{tab:h2g_results}
    \vspace{-0.5cm}
\end{table}

\section{Additional Visualizations}
\label{sec:supp_vis}

\begin{figure*}[!t]
     \centering
    \includegraphics[width=0.8\textwidth]{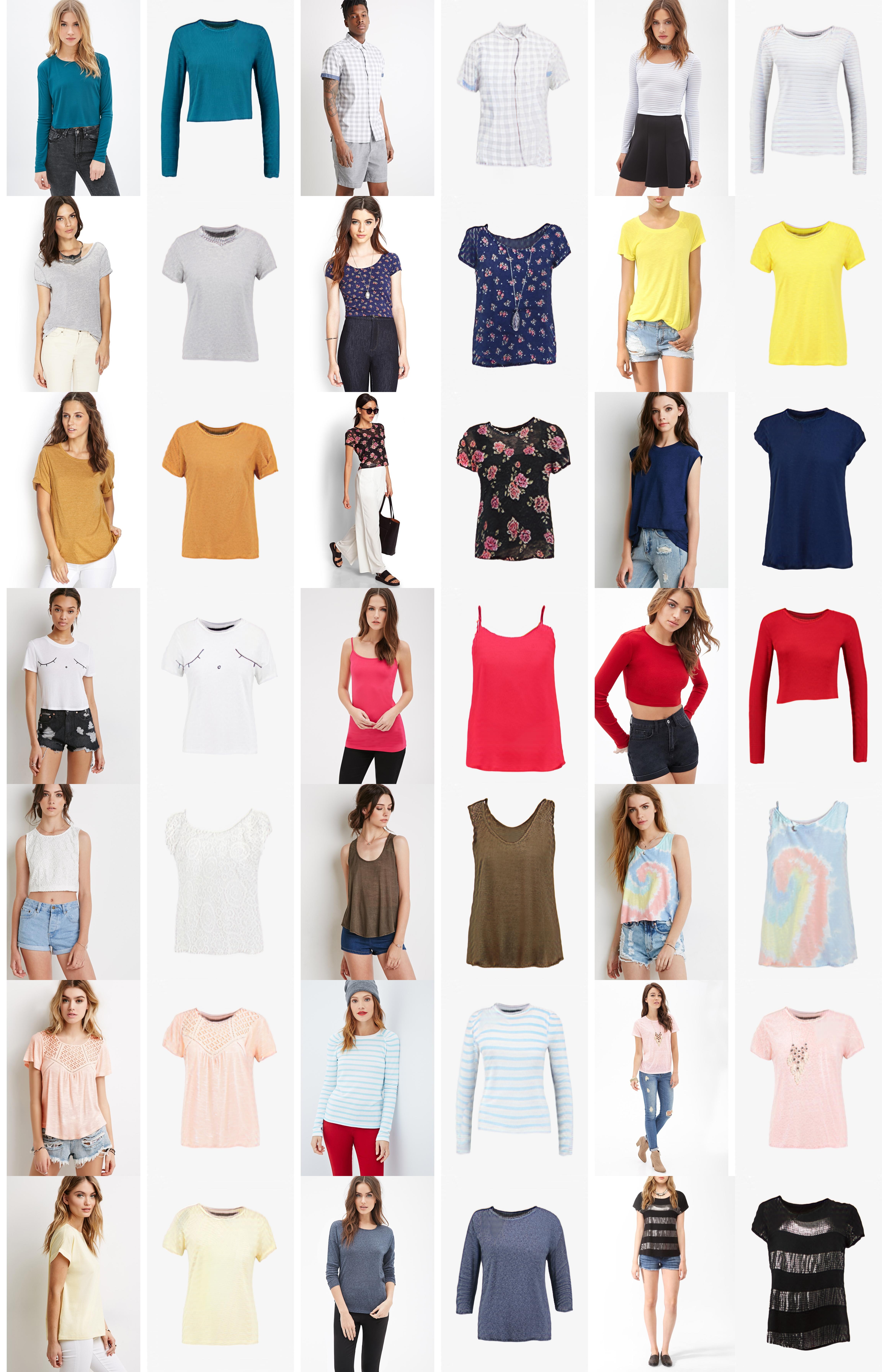}
    \caption{Visualized examples of the (human, synthetic garment) pairs on our proposed H2G-UH.}
    \label{fig:more_vis_h2g_uh}
\end{figure*}

\begin{figure*}[!t]
     \centering
    \includegraphics[width=0.8\textwidth]{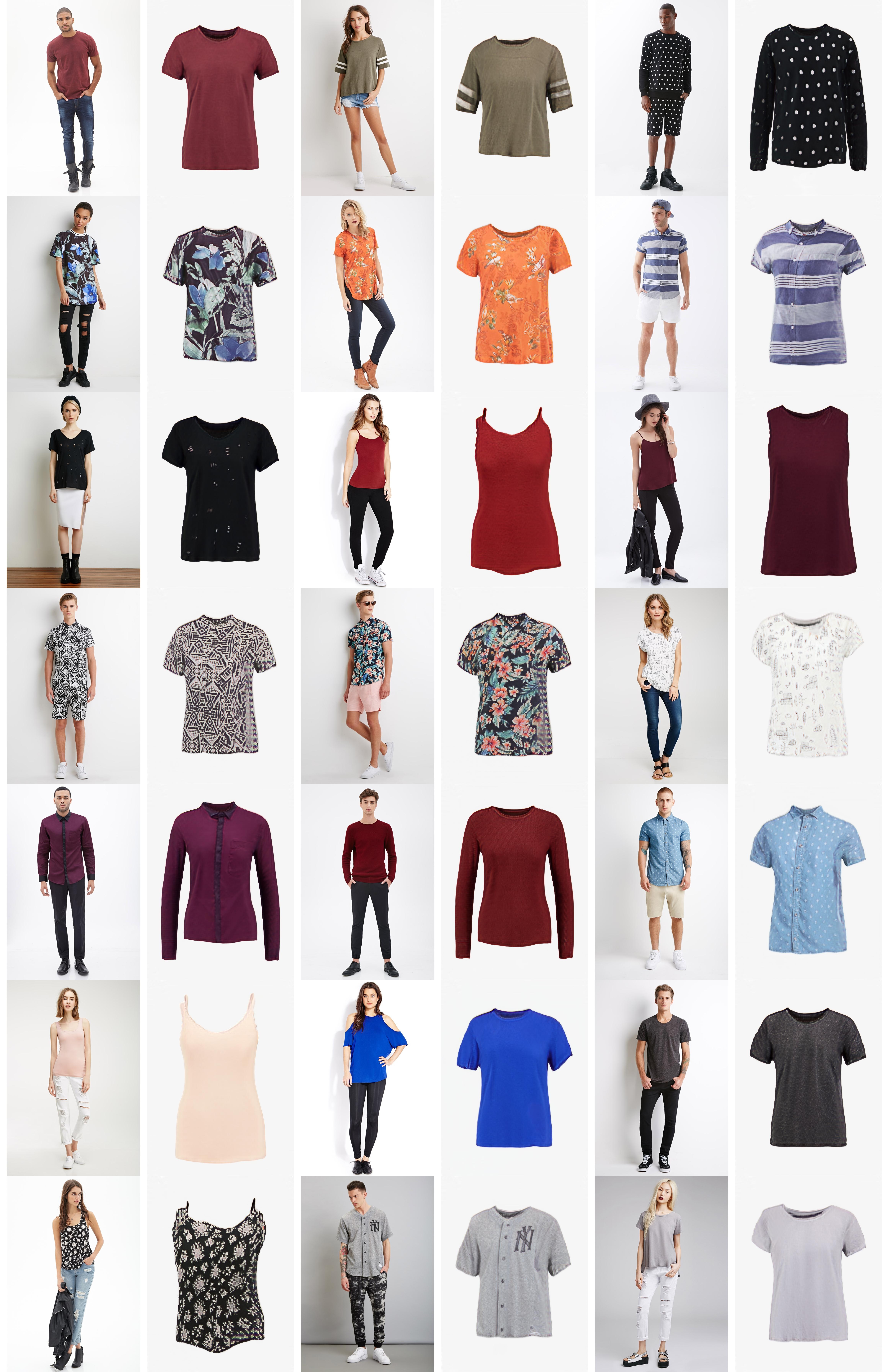}
    \caption{Visualized examples of the (human, synthetic garment) pairs on our proposed H2G-FH.}
    \label{fig:more_vis_h2g_fh}
\end{figure*}

Figures \ref{fig:more_vis_h2g_uh} and \ref{fig:more_vis_h2g_fh} showcase exemplars from our synthesized datasets H2G-UH and H2G-FH, respectively. We also report quantitative results in Table \ref{tab:h2g_results} to evaluate our human-to-garment model on VITON-HD and DressCode-Upper. The generated garment images in Figures \ref{fig:more_vis_h2g_uh} and \ref{fig:more_vis_h2g_fh} closely mimic the product view of the clothing items, accurately capturing both the shape and texture of the original garments worn by the individuals. This approach to creating synthetic training data for the virtual try-on task is both cost-effective and data-efficient, highlighting the benefits of our proposed human-to-garment model.

\begin{figure*}[!t]
     \centering
    \includegraphics[width=\textwidth]{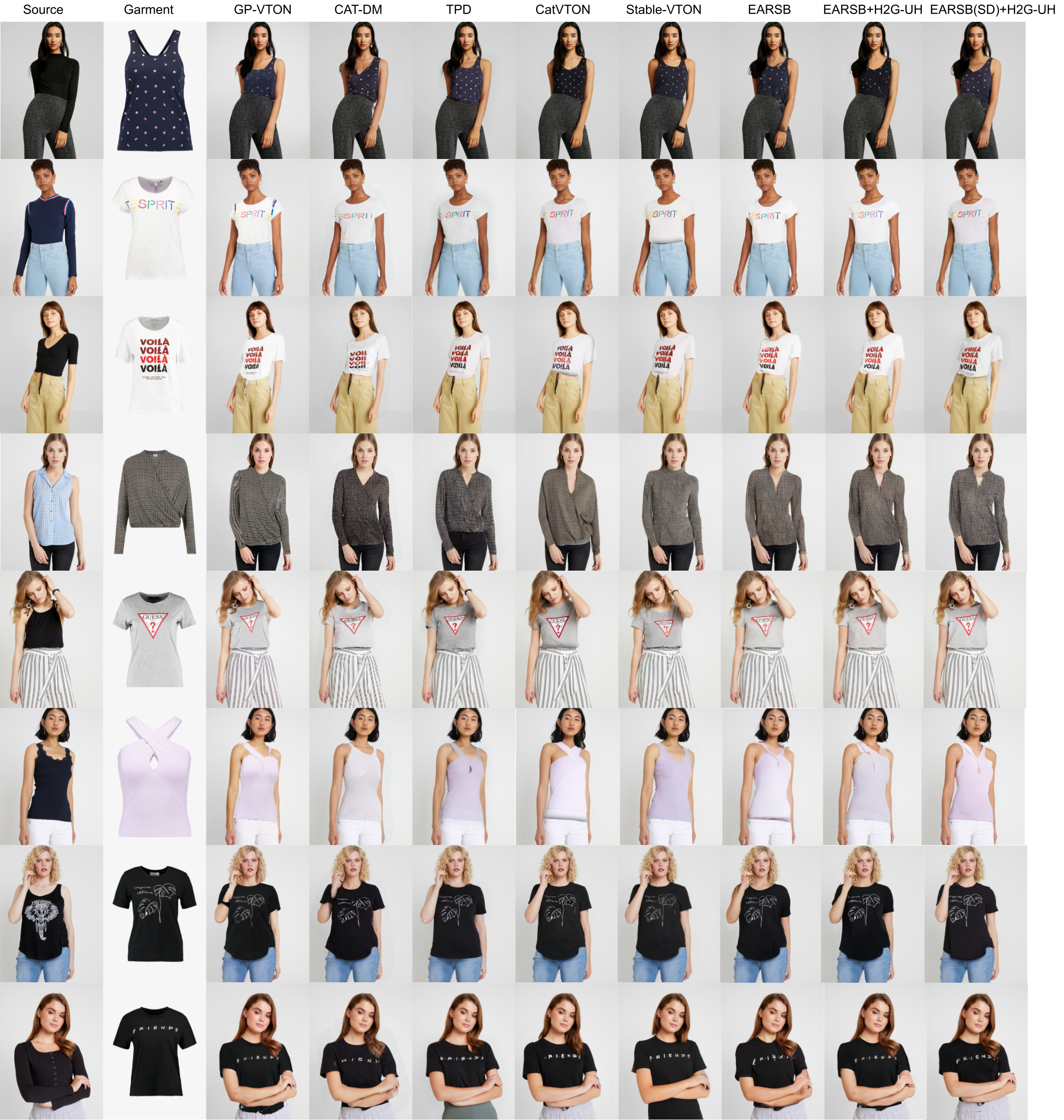}
    \caption{Visualized examples on VITON-HD. Our \Ours{} and \Ours{}+H2G-UH better recovers the intricate textures in the garment.}
    \label{fig:more_vis_viton}
\end{figure*}

\begin{figure*}[!t]
     \centering
    \includegraphics[width=\textwidth]{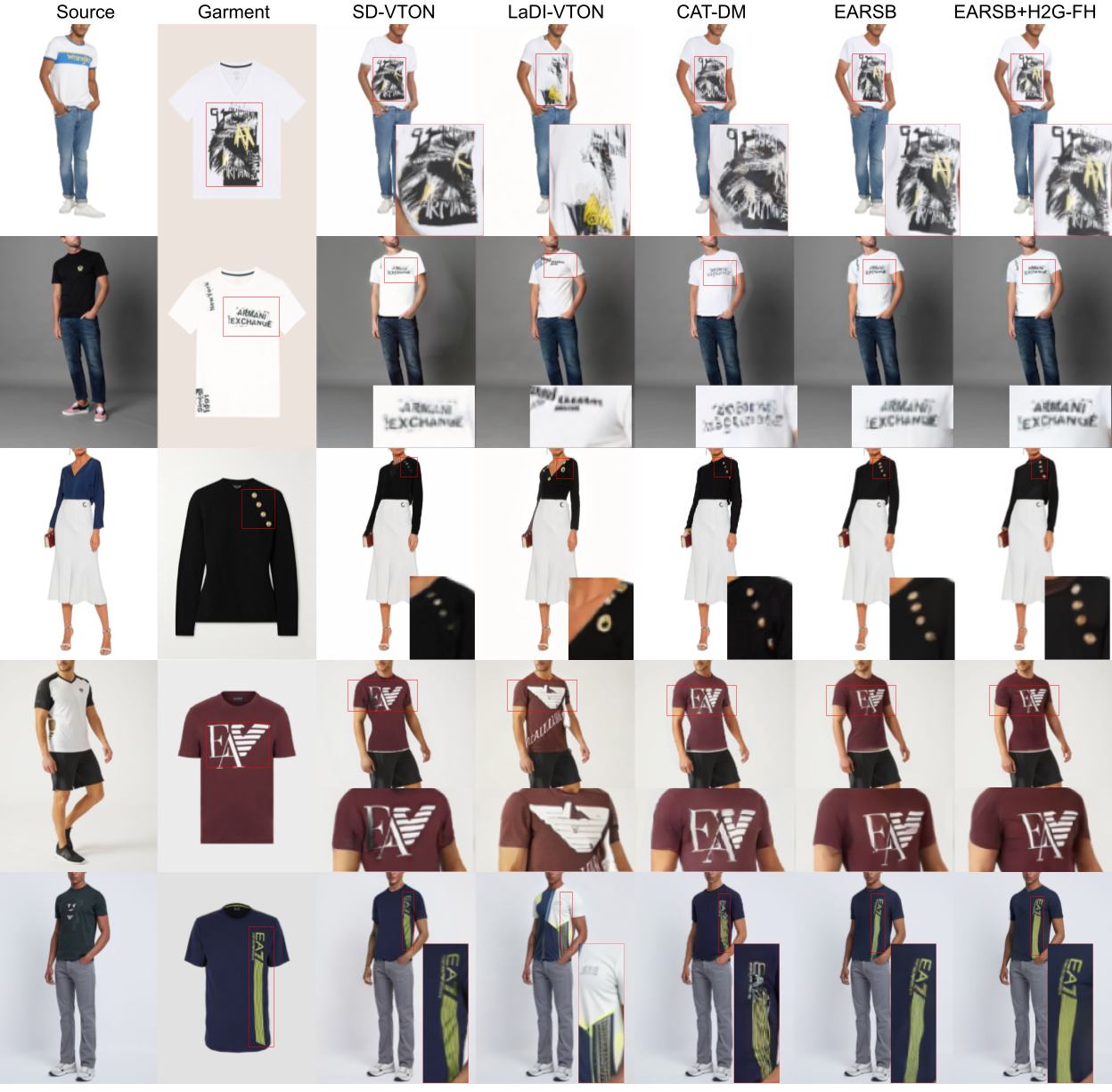}
    \caption{Visualized examples on DressCode-Upper.  Our \Ours{} and \Ours{}+H2G-UH better reconstructs the texts and graphics in the garment.}
    \label{fig:more_vis_dc}
\end{figure*}

Figures \ref{fig:more_vis_viton} and \ref{fig:more_vis_dc} give visualized results of the proposed \Ours{} and \Ours{}+H2G-UH. In contrast to previous approaches, \Ours{} specifically targets and enhances low-quality regions in GAN-generated images, which typically correspond to texture-rich areas. This targeted improvement is evident in the last row of Fig. \ref{fig:more_vis_viton}, where \Ours{} more accurately reconstructs text \emph{freinds}, and in the third row, where it successfully generates four side buttons. Furthermore, the incorporation of our synthetic dataset H2G-UH with \Ours{} leads to even more refined details in the generated images, demonstrating the synergistic effect of our combined approach.

\section{Ethics}
We acknowledge several potential ethical considerations of our work on virtual try-on:
\begin{itemize}
    \item Bias and representation: We strive for diversity in our training data to ensure the model performs equitably across different body types, skin tones, and ethnicities. However, biases may still exist, and further work is needed to assess and mitigate these.
    \item Misuse potential: While intended for benign purposes, this technology could potentially be misused to create misleading or non-consensual images. We strongly condemn such uses and will explore safeguards against misuse in future work.
    %\item Environmental impact: While our method aims to reduce environmental impact from returns, the computational resources required for training and running these models have their own carbon footprint. We are committed to optimizing our models for efficiency.
\end{itemize}
We believe the potential benefits of this technology outweigh the risks, but we remain vigilant about these ethical considerations and are committed to addressing them as our research progresses.

\clearpage
\clearpage

\clearpage

\clearpage

\end{document}